\documentclass[runningheads]{llncs}

\usepackage{accv}

\usepackage{accvabbrv}
\usepackage{graphicx}
\usepackage{booktabs}
\usepackage{amsmath}
\usepackage{amssymb}
\usepackage[accsupp]{axessibility}

\usepackage{hyperref}
\usepackage{orcidlink}

\begin{document}

\title{Quantifying Annotation Agreement in Object Detection with Optimal Transport}

\titlerunning{Optimal Transport for Annotation Agreement}

\author{Kirill Prokhorov\inst{1}\orcidlink{0000-0001-5915-6614} \and
Alexandr A. Kalinin\inst{2}\orcidlink{0000-0003-4563-3226}}

\authorrunning{K.~Prokhorov et al.}

\institute{Openface.io, Vilnius, Lithuania\\
\email{prokhorov@openface.me}
\and
Broad Institute of MIT and Harvard, Cambridge, MA, USA}

\maketitle

\begin{abstract}
Reliable validation of machine learning models is a crucial step in designing robust and trustworthy systems. Yet validation relies on annotated test sets treated as ground truth, even though annotations can be noisy or ambiguous. In object detection, sources may disagree on object presence, box location, or class label. Standard agreement metrics usually binarize box overlap before computing a summary score, obscuring the degree and structure of disagreement between annotation sets. Here, we study an Optimal Transport (OT)-based metric for measuring agreement between object detection annotations. Instead of binarizing bounding box overlap, this metric assigns graded costs to matched and unmatched labeled boxes. Through theoretical analysis and experiments on real and synthetically perturbed annotations, we show that OT captures graded changes in agreement and distinguishes annotation pairs that thresholded metrics score identically. Our results suggest that OT provides a richer measure of annotation agreement in object detection, especially for domains with many small or ambiguous objects. Such agreement estimates can support more reliable validation when labels are noisy or conflicting.
\end{abstract}

\section{Introduction}
\label{sec:intro}

Modern machine learning systems rely on annotated data for training and
validation, typically treating a fixed test set as a \emph{gold standard}.
Yet a single noise-free ground truth may not exist: multiple plausible
annotations can arise from task ambiguity, limited visual evidence, or
differences in expert judgment. This is especially relevant in object
detection, where annotators may disagree on object presence, spatial
extent, or class labels~\cite{feng2021labels,murrugarra2022can,drawing_same_bounding_box_twice}.
Such disagreement contains information about the reliability of the
reference data. A prediction that differs from one annotation may remain
consistent with another plausible interpretation, so performance against
a single reference can obscure annotation uncertainty. Measuring
agreement is therefore important for assessing both dataset quality and
the reliability of model validation.

Agreement in object detection is commonly assessed by matching boxes with an Intersection-over-Union (IoU) threshold and then computing a summary statistic such as F1 or Krippendorff's $\alpha$~\cite{krippendorff_alpha_reliability,krippendorf_obj_det}. This approach is simple and useful, but the thresholding step discards spatial information. After overlap is converted into a binary match decision, the resulting score may no longer reflect how close the boxes are, how severe the mismatch is, or how unmatched boxes are arranged. Thus, annotation pairs with different spatial structure can receive the same agreement score.

In this work, we study an Optimal Transport (OT)-based metric for measuring agreement between object detection annotations. The formulation builds on recent OT-based work for object detection evaluation~\cite{otani2022optimal}, but shifts the focus from comparing model predictions with a reference to comparing annotation sets themselves. Each annotation is treated as a set of labeled bounding boxes, and agreement is computed through a transport problem whose cost combines localization, label consistency, and penalties for unmatched boxes.

Our main contributions are:
\begin{itemize}
  \item We adapt an OT-based detection metric to quantify agreement
  between object detection annotations, jointly accounting for
  localization, class labels, and missing or extra boxes.
  \item We analyze its bounds and special cases, clarifying how object
  count, localization cost, and matching penalties affect the score.
  \item We evaluate OT on real and synthetically perturbed annotations
  from multiple domains, identifying cases where it captures distinctions
  collapsed by threshold-based agreement metrics.
\end{itemize}

Together, these results support OT-based agreement as a practical tool for assessing object detection annotations and for making model validation more sensitive to noisy or conflicting labels. The source code is publicly available at \url{https://github.com/openface-io/ot-agreement}.

\section{Related Work}
\label{sec:related}

\subsection{Uncertainty and disagreement in object detection annotations}

Annotation uncertainty has been addressed by modeling label noise or
learning from repeated annotations~\cite{drawing_same_bounding_box_twice,learn2agree},
using richer targets such as label distributions~\cite{geng_ldl,joint_acne_grading_ldl,acne_grading_lds},
representing localization uncertainty probabilistically~\cite{gaussian_bboxes_prob_iou},
and aggregating annotations into a latent ground truth~\cite{learn2agree,annotator_agreement_empirical_study,staple}.
Other work characterizes model performance under contradictory
labels~\cite{label_convergence}. Our goal is complementary: we measure
pairwise agreement and discrepancy directly between object detection
annotations, either to characterize annotation quality or as a component
of multi-reference validation.

\subsection{Agreement and evaluation metrics for object detection}

Object detection commonly uses IoU-threshold matching followed by metrics
such as precision, recall, F1, or mAP. Similar matching is used when
adapting agreement statistics such as Krippendorff's $\alpha$ to object
detection~\cite{krippendorf_obj_det,krippendorff_alpha_reliability}.
Although interpretable, hard thresholds discard information about the
structure and severity of spatial disagreement.

Methods such as the discrepancy ratio~\cite{discrepancy_ratio} compare
model predictions with multiple human annotations and therefore depend
on an underlying pairwise agreement measure. A more expressive pairwise
measure can improve how such methods characterize annotation variability
and model performance relative to it.

\subsection{Optimal transport for detection evaluation}

Optimal Transport provides a natural framework for comparing structured sets by computing a minimum-cost assignment between elements. With dummy elements or related unbalanced formulations, it can also account for unmatched or poorly matched elements. More broadly, transport-style objectives have been used to compare distributions, align structured representations, and solve set-matching problems~\cite{ota,de2023unbalanced,sarlin2020superglue}.

In object detection, Otani et al.~\cite{otani2022optimal} proposed an OT-based evaluation metric, \emph{Optimal Correction Cost} (OC-cost), building on earlier work~\cite{ozdemir2010performance,ota}. OC-cost compares detector predictions with ground-truth boxes and penalizes localization errors, label mismatches, false positives, and false negatives within a single transport problem. This produces a fine-grained measure of detection quality and addresses limitations of standard mAP-style evaluation.

Our work uses the same general OT perspective but changes the object of comparison. Rather than evaluating detector predictions against a fixed ground truth with confidence scores, we compare annotation sets directly. This shift removes the need to treat either annotation as error-free and allows OT to serve as a measure of annotation agreement or discrepancy. In this setting, the transport cost captures spatial proximity, label consistency, and missing or extra boxes, providing a graded alternative to thresholded agreement metrics.

\section{Method}
\label{sec:method}

\subsection{Optimal Transport score formulation}

Our formulation follows Otani et al.~\cite{otani2022optimal}, who use OT
to compare detections with ground truth. In the multi-annotator setting,
we compare only bounding boxes and class labels, without detector
confidence scores.

Let $\mathcal{A}=\{b_1^\mathcal{A},\dots,b_M^\mathcal{A}\}$ and
$\mathcal{B}=\{b_1^\mathcal{B},\dots,b_N^\mathcal{B}\}$ be two sets of
$M$ and $N$ labeled bounding boxes. Let $l_i^\mathcal{A}$ and
$l_j^\mathcal{B}$ denote their corresponding class labels. The pairwise
OT cost is
\begin{equation}
    c_{ij} = \lambda\, c_{\text{loc}}(b_i^\mathcal{A}, b_j^\mathcal{B}) + (1 - \lambda)\, c_{\text{lbl}}(l_i^\mathcal{A}, l_j^\mathcal{B}),
\label{eq:ot_cost}
\end{equation}
where $\lambda\in[0,1]$ is a trade-off hyperparameter that balances localization and label costs. The localization cost is defined using Generalized IoU (GIoU)~\cite{giou}:
\begin{equation}
    c_{\text{loc}}(b_i^\mathcal{A}, b_j^\mathcal{B}) = \frac{1 - GIoU(b_i^\mathcal{A}, b_j^\mathcal{B})}{2}, \quad c_{\text{loc}}(b_i^\mathcal{A}, b_j^\mathcal{B})\in[0, 1].
    \label{eq:spat_dist}
\end{equation}
The properties of GIoU are well suited to our task, as they yield meaningful distances even for non-overlapping boxes. The label cost is given by
\begin{equation}
    c_{\text{lbl}}(l_i^\mathcal{A}, l_j^\mathcal{B}) = 
    \begin{cases}
        0, & \text{if } l_i^\mathcal{A} = l_j^\mathcal{B}, \\
        1, & \text{if } l_i^\mathcal{A} \neq l_j^\mathcal{B}.
    \end{cases}
    \label{eq:ldb_dist}
\end{equation}
We then construct the cost matrix:
\begin{equation}
C = 
\begin{pmatrix}
    c_{11} & \cdots & c_{1N} & \beta \\
    \vdots & \ddots & \vdots & \vdots \\
    c_{M1} & \cdots & c_{MN} & \beta \\
    \beta  & \cdots & \beta  & \beta
\end{pmatrix},
\end{equation}
where the last row and column correspond to dummy boxes with constant
cost $\beta\in[0,1]$. A real box may match a dummy when no suitable real
counterpart exists, representing a missing or extra object.

Within the OT framework, we formulate the task as mass transportation from one set to the other. We specify the source and target marginal distributions as weight vectors \cite{peyre2019computational}:
\begin{equation}
    \mathbf{a} = [\underbrace{1, 1, \dots, 1}_{M},\, N], \quad 
    \mathbf{b} = [\underbrace{1, 1, \dots, 1}_{N},\, M],
\end{equation}
where the last entry in each vector corresponds to the dummy box. Unit capacity is assigned to each real box and an aggregate capacity to the dummy boxes. Each real box carries unit mass; because the marginals are integral,
the transport problem admits an integral optimal assignment.

The optimal transport plan $\pi$ is then obtained by solving the optimization problem:
\begin{equation}
\begin{aligned}
    \pi^* &= \underset{\pi}{\operatorname{argmin}} \sum_{i=1}^{M+1}\sum_{j=1}^{N+1} c_{ij}\,\pi_{ij}, \\
    \text{subject to } &\sum_{j=1}^{N+1}\pi_{ij} = a_i, \quad \sum_{i=1}^{M+1}\pi_{ij} = b_j \quad \forall i,j, \\
                       &\sum_{i=1}^{M+1} a_i = \sum_{j=1}^{N+1} b_j, \\
                       &\pi_{ij} \geq 0 \quad \forall i,j.
\end{aligned}
\end{equation}

After computing the optimal transport plan $\pi^*$ (e.g. using solvers implemented in~\cite{flamary2021pot}), we ignore dummy-to-dummy mass by setting $\pi^*_{M+1,N+1}=0$. The final OT cost is then defined as
\begin{equation}
    \tilde{c} = \frac{1}{\sum_{i=1}^{M+1}\sum_{j=1}^{N+1}\pi^*_{ij}} \sum_{i=1}^{M+1}\sum_{j=1}^{N+1} c_{ij}\,\pi_{ij}^*.
    \label{eq:final_cost}
\end{equation}
Let $k$ denote the number of real-real matches. We can then write the normalization factor as
\[
Z = \sum_{i=1}^{M+1}\sum_{j=1}^{N+1}\pi^*_{ij} =\underbrace{k}_\text{real-real}+\underbrace{(M-k)}_{\text{real}\ \mathcal{B}\rightarrow \text{dummy}\ \mathcal{A}}+\underbrace{(N-k)}_{\text{real}\ \mathcal{A}\rightarrow\text{dummy}\ \mathcal{B}}=M+N-k.
\]

We constrain $\beta\leq1$, matching the range of the pairwise cost in
Eq.~\ref{eq:ot_cost}. Here $\beta$ is the cost of a real--dummy match.
Because dummy matching provides a feasible fallback, the optimal
normalized OT cost is at most $\beta$. Thus the OT cost lies in
$[0,\beta]$, while the corresponding score
$OT_{\rm score}=1-OT_{\rm cost}$ lies in $[1-\beta,1]$.

The raw OT score measures pairwise similarity but does not account
for agreement expected by chance. We therefore additionally define
a dataset-level, chance-corrected agreement coefficient using the
$\alpha=1-D_o/D_e$ form~\cite{braylan2022measuring}:

\begin{equation}
\alpha_{\rm OT}=1-\widehat D_o/\widehat D_e,
\label{eq:chance_corr}
\end{equation}
where $\widehat D_o$
averages OT costs between annotations of the same image and
$\widehat D_e$ is estimated from Monte Carlo-sampled cross-image pairs.
Thus $\alpha_{\rm OT}=1$, $0$, and $<0$ denote perfect, chance-level,
and below-chance agreement, respectively.

\subsection{Special cases and theoretical boundaries}

We next study how the OT score behaves under simplified settings that make the transport plan interpretable. For arbitrary annotations, the optimal plan depends on the number of boxes, labels, and hyperparameters and must generally be computed numerically. To obtain analytical insight, we consider special cases in which the matching structure is known or can be approximated. These cases clarify how the score depends on object count, spatial mismatch, and the dummy-matching penalty.

\subsubsection{Case: dummy matching only for unequal box counts}
Consider two sets of unlabeled ($c_{\text{lbl}}=0$) annotations
$\mathcal{A}=\{b_1^\mathcal{A},\dots,b_M^\mathcal{A}\}$ and
$\mathcal{B}=\{b_1^\mathcal{B},\dots,b_N^\mathcal{B}\}$.
If $\lambda\leq\beta\leq1$, real--real matches are no more costly
than dummy matches. Thus dummy matches occur only when the annotations
contain different numbers of boxes.

If all real boxes have to match, then $Z = \max(M, N)$ and upper and lower bounds for OT score have the form:

\begin{equation}
1-\frac{1}{Z}\left( \underbrace{\beta |M-N|}_{\text{diff. num. of boxes}} + \underbrace{\lambda \min(M,N)}_{\text{max spatial mismatch}} \right) \leq 1-OT \leq 1-\frac{1}{Z} \underbrace{\beta|M-N|}_{\text{diff. num. of boxes}}
\label{eq:spat_bounds}
\end{equation}

The minimum possible value depends on the box-count mismatch and the
maximum spatial mismatch ($c_{\text{loc}}=1$). The maximum is attained
when matched boxes are perfectly aligned ($c_{\text{loc}}=0$).
If $\beta<\lambda$, all boxes may match to dummies in the worst case,
and the lower bound becomes $1-\beta$.

We can expand Eq.~\ref{eq:spat_bounds} for two cases:
\begin{equation}
\begin{cases}
  1-\left(\beta (1-\frac{N}{M})+\lambda \frac{N}{M} \right) \leq 1-\text{OT} \leq 1-\beta(1-\frac{N}{M}), & \text{if } N \leq M \\[10pt]
  1-\left(\beta (1-\frac{M}{N})+\lambda \frac{M}{N} \right) \leq 1-\text{OT} \leq 1-\beta(1-\frac{M}{N}), & \text{if } N > M
\end{cases}
\label{eq:boundaries}
\end{equation}

\begin{figure}[tb]
  \centering
  \includegraphics[height=6.5cm]{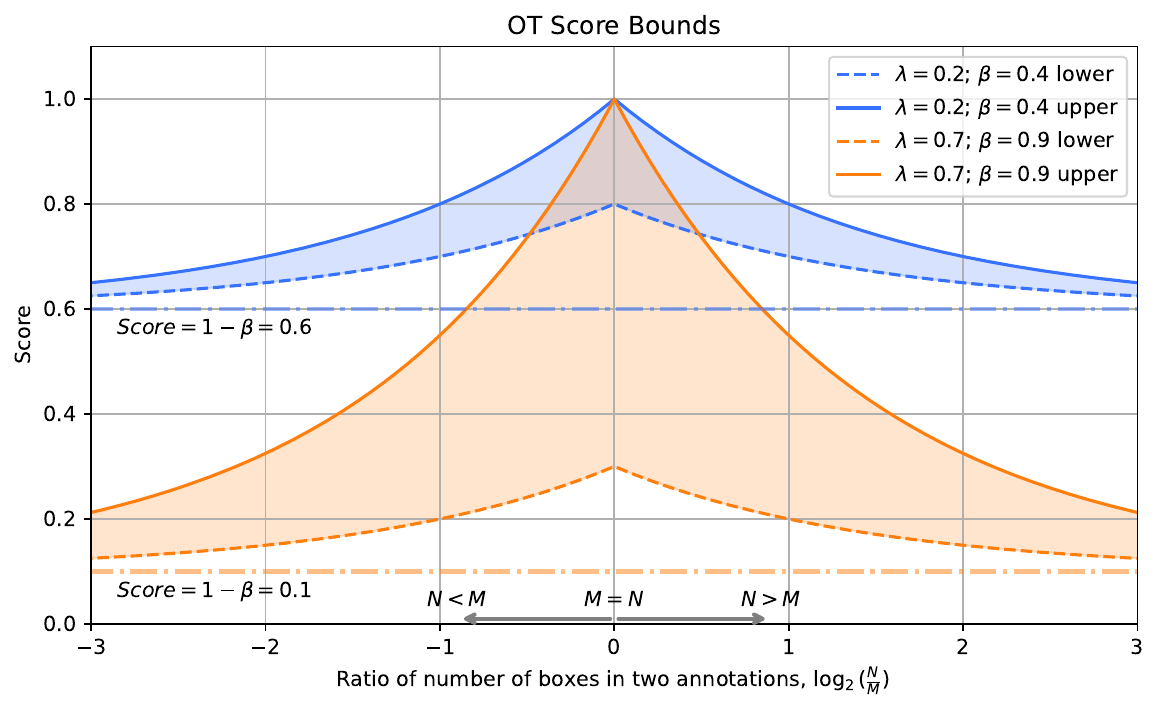}
  \caption{
    OT score bounds and corresponding range of values as a function of the ratio between the numbers of boxes in two annotations. When the numbers of boxes coincide, the range equals $\lambda$, while $\beta$ controls the maximum (minimum) values of the cost (score). All curves asymptotically converge as $\log_{2}\!\left(\tfrac{N}{M}\right) \to\pm\infty$. The sensitivity of the optimal transport metric to spatial localization gives rise to the full range of possible values.
}
  \label{fig:ot_vs_diff}
\end{figure}

It is convenient to plot Eq.~\ref{eq:boundaries} against $\log_2\!\left(\frac{N}{M }\right)$ so it becomes symmetric. Fig.~\ref{fig:ot_vs_diff} shows boundaries and range of possible values for two cases with different $\lambda$ and $\beta$.
Note that when the difference in the number of boxes is small, the range of metric values is large and equal to $\lambda$ when $\log_2\!\left(\frac{N}{M }\right)=0$. 
When the number of boxes differs significantly, the range is small, because the metric is poor in any case regardless of box locations. As $\log_2\!\left(\frac{N}{M }\right)\to\pm\infty$ both upper and lower bounds approach $1-\beta$ and the range of possible values shrinks to zero.

\subsubsection{Case: dependence on $\beta$ under the noisy-box assumption.}
Consider two sets of unlabeled boxes
$\mathcal{A}=\{b_1^\mathcal{A},\dots,b_M^\mathcal{A}\}$ and
$\mathcal{B}=\{b_1^\mathcal{B},\dots,b_M^\mathcal{B}\}$.
We analyze how $\beta$ affects whether boxes are matched to real boxes or to dummies.

Because this example includes only the spatial component of the OT cost, we set $\lambda=1$ and assume for simplicity that corresponding boxes have the same dimensions $(w,h)$. For each matched pair, assume that the box centers differ by noise:

\begin{equation}
    \begin{cases}
  x_{c2} = x_{c1} + \varepsilon_x \\
  y_{c2} = y_{c1} + \varepsilon_y
  \end{cases}
\end{equation}

In order to find GIoU for spatial cost term in Eq.~\ref{eq:spat_dist}, we have to define intersection
$I = \max(0, w-|\varepsilon_x|)\cdot \max(0, h-|\varepsilon_y|)$, union $U = 2wh-I$ and 
the area of the smallest enclosing box for two boxes $|C| = (w+|\varepsilon_x|)(h+|\varepsilon_y|)$.
This gives the following expression for GIoU:
\begin{equation}
  GIoU = IoU - \frac{|C \setminus U|}{|C|} = \frac{I}{2wh-I} - 
  \frac{(w+|\varepsilon_x|)(h+|\varepsilon_y|) - (2wh - I)}{(w+|\varepsilon_x|)(h+|\varepsilon_y|)}.
  \label{eq:giou}
\end{equation}

A real--real match is preferred to a dummy match when $c_{ij}<\beta$, while $c_{ij}>\beta$ favors matching to the dummy. Therefore, the probability of a real--real match is
\begin{equation}
  P(c_{ij}<\beta) = F_c(\beta) = \int\limits_{0}^{\beta} f_c(t) \, dt = 
  \iint\limits_{c(\xi, \eta) < \beta}f_{|\varepsilon_x|, |\varepsilon_y|}(\xi, \eta) \, d\xi d\eta
  \label{eq:prob}
\end{equation}
where $f_{|\varepsilon_x|, |\varepsilon_y|}(|\varepsilon_x|, |\varepsilon_y|)$ is the joint PDF for the noise components $|\varepsilon_x|$ and $|\varepsilon_y|$.

To evaluate this probability, we need the distribution of the cost $c$.
Since only the noise distribution is assumed known, we instead integrate over the joint density $f_{|\varepsilon_x|,|\varepsilon_y|}(\xi,\eta)$.
The remaining task is therefore to determine the boundary
$c(|\varepsilon_x|,|\varepsilon_y|)=\beta$, which defines the integration domain in Eq.~\ref{eq:prob}.

To derive this boundary from Eq.~\ref{eq:giou}, we consider separately the cases $I>0$ and $I=0$. The former corresponds to
$|\varepsilon_x|<w$ and $|\varepsilon_y|<h$, whereas the latter occurs when either $|\varepsilon_x|\geq w$ or $|\varepsilon_y|\geq h$.
Introducing the normalized displacements
$u=|\varepsilon_x|/w$ and $\vartheta=|\varepsilon_y|/h$, and using Eq.~\ref{eq:spat_dist}, the condition
$c(|\varepsilon_x|,|\varepsilon_y|)=\beta$ is equivalent to
$GIoU=1-2\beta$. For the case $I=0$, this gives:



\begin{equation}
  GIoU_{I=0} = -\frac{(1+u)(1+\vartheta) - 2}{(1+u)(1+\vartheta)}= 1-2\beta.
  \label{eq:giou_0}
\end{equation} 
Similarly, for $I\neq 0$:

\begin{equation}
  GIoU_{I\neq 0} = \frac{(1-u)(1-\vartheta)}{2-(1-u)(1-\vartheta)}-
  \frac{(1+u)(1+\vartheta)-2+(1-u)(1-\vartheta)}{(1+u)(1+\vartheta)} = 1-2\beta.
  \label{eq:giou_not_0}
\end{equation}
From Eq.~\ref{eq:giou_0} it is easy to find $\vartheta_{I=0}(u)$; Eq.~\ref{eq:giou_not_0} can be rewritten as a quadratic equation 
$A(u)\vartheta^2+B(u)\vartheta+C(u)=0$ and solved in order to obtain $\vartheta_{I\neq0}(u)$. 
These functions determine the integration limits
$a_1(u),a_2(u),a_3(u)$ for the three regions. Assuming independent noise components, Eq.~\ref{eq:prob} becomes



\begin{multline}
  F_c(\beta) = \int\limits_{0}^{1} f_{u}(\xi) F_{\vartheta}(a_1(\xi)) \, d\xi  +
  \int\limits_{0}^{1} f_{u}(\xi)\left[F_{\vartheta}(a_2(\xi)) - F_{\vartheta}(1)\right] \, d\xi  + \\
  + \int\limits_{1}^{\infty}f_{u}(\xi)F_{\vartheta} (a_3(\xi)) \, d\xi 
  \label{eq:cdf_final}
\end{multline}
Detailed derivation can be found in the supplementary material.
Given a PDF for the noise distribution, these integrals can be evaluated numerically.

\begin{figure}[tb]
  \centering
  \begin{subfigure}{\linewidth}
  \centering
    \includegraphics[width=0.9\linewidth]{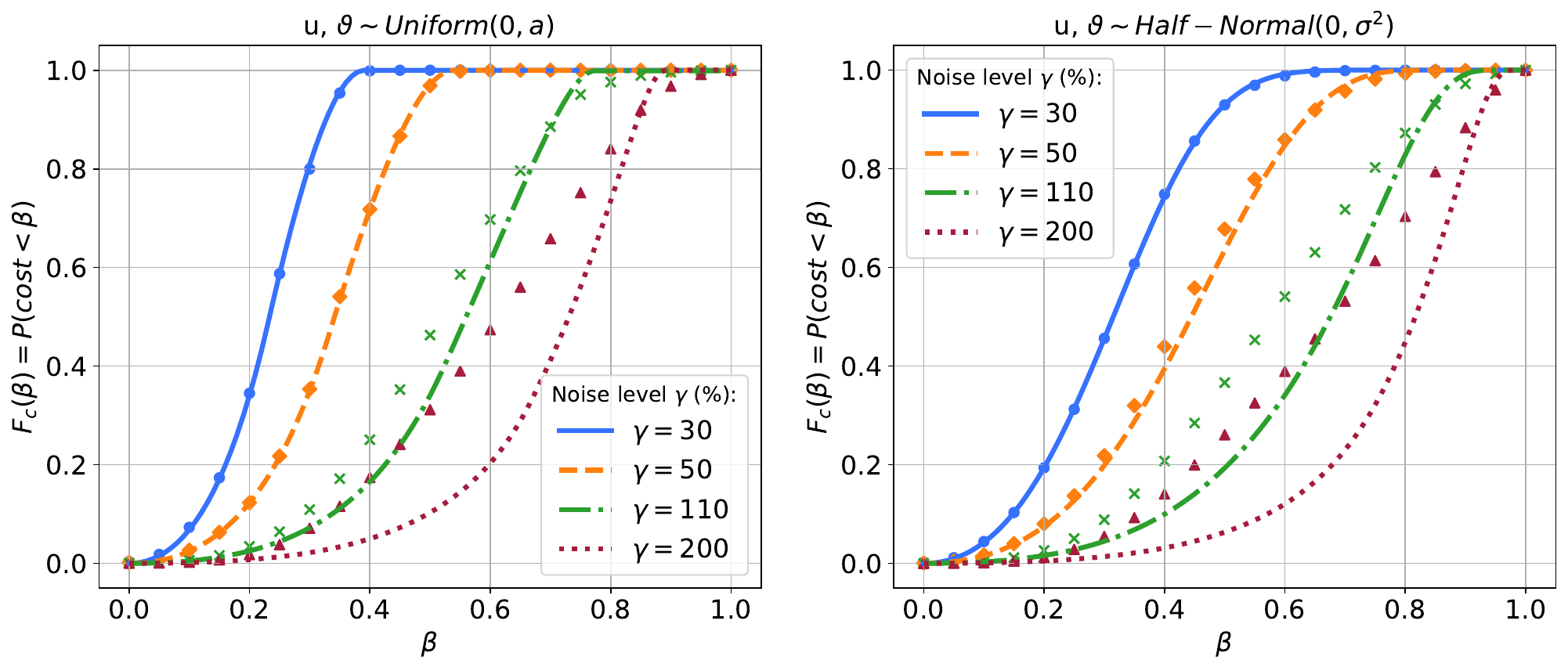}
    \caption{}
    \label{fig:cdf}
  \end{subfigure}
  \vfill
  \begin{subfigure}{\linewidth}
  \centering
    \includegraphics[width=0.9\linewidth]{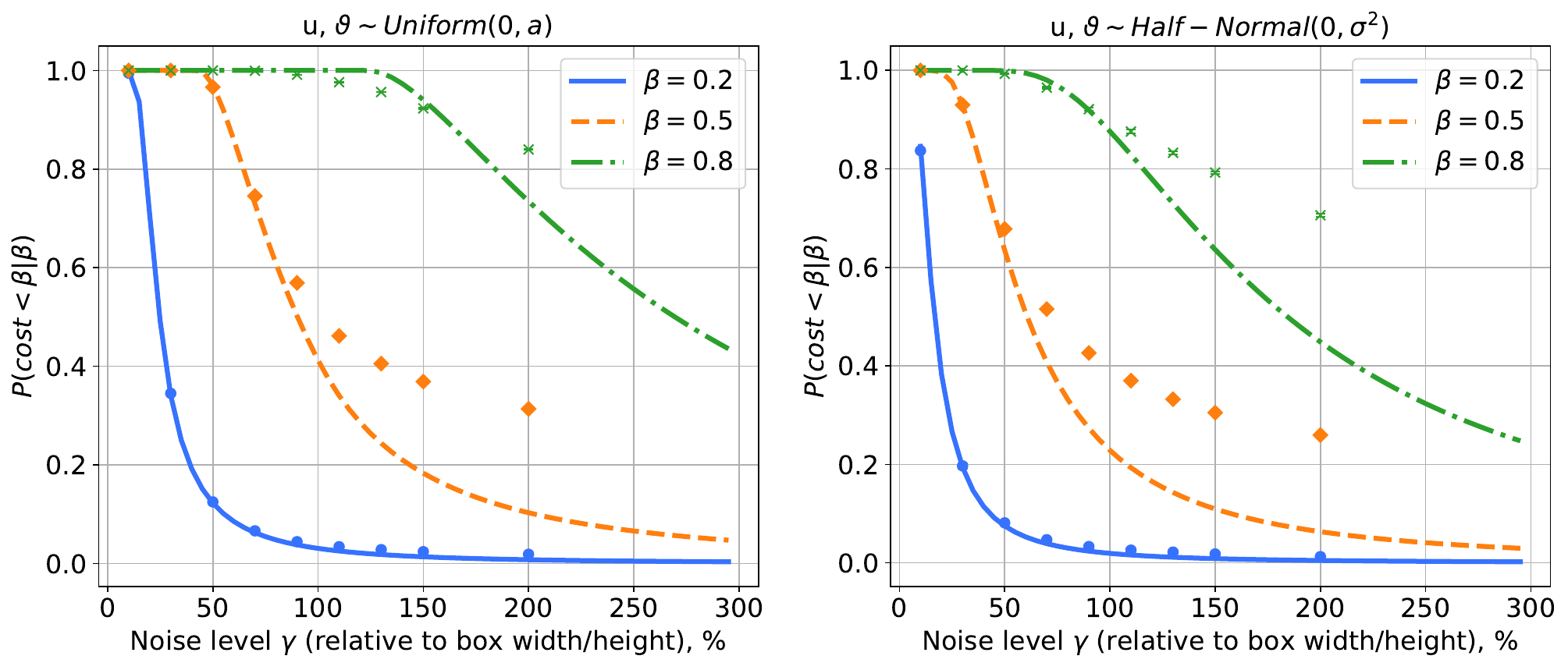}
    \caption{}
    \label{fig:prob_vs_rel}
  \end{subfigure}
\caption{Probability of real--real matching under the noisy-box
assumption. \textbf{a)} Dependence on $\beta$ for several noise levels $\gamma$. \textbf{b)} Dependence on $\gamma$ for fixed $\beta$. Theoretical and empirical results diverge at higher noise levels as matches break and the noisy-box assumption becomes less accurate.}
  \label{fig:example4-6}
\end{figure}

Fig.~\ref{fig:cdf} shows the theoretical CDF of the box-matching cost as a function of $\beta$ under Uniform and Half-Normal perturbations, together with empirical estimates on ACNE04~\cite{joint_acne_grading_ldl}. For visualization, we use the annotation set from~\cite{acneai}, which contains substantially more objects than the original annotation. Each curve corresponds to a different noise level $\gamma$.
As $\beta$ increases, a perturbed box is more likely to match a real box rather than the dummy box, because $\beta$ sets the penalty for unmatched boxes. Higher noise levels shift the curves toward larger $\beta$: when boxes are strongly perturbed, real-real matches require a larger dummy penalty to remain preferable.
The theoretical and empirical curves agree well for small perturbations, where each perturbed box remains close to its original counterpart. The discrepancy grows at higher noise levels. In this regime, a perturbed box may become closer to a different object than to the box from which it was generated, violating the noisy-box assumption used in the derivation. This deviation is informative: it indicates that the two annotations are no longer related by simple localization noise, but differ more substantially in spatial structure.

Fig.~\ref{fig:prob_vs_rel} shows the same effect from another perspective, plotting the probability of real-box matching as a function of noise level $\gamma$ for fixed values of $\beta$. Larger $\beta$ values allow boxes with higher spatial cost to remain matched to real boxes rather than to the dummy box, so the noisy-box approximation remains valid over a wider range of perturbations. For small $\beta$, such as $\beta=0.2$, boxes are sent to the dummy box even under modest shifts, and the assumed one-to-one noisy correspondence breaks down earlier.






\section{Experiments}
\label{sec:experiments}

\subsection{Datasets}
\label{sec:datasets}

\begin{itemize}
\item \textbf{ACNE04}. We use ACNE04 because two independent annotation sets are available for the same domain, enabling direct comparison of real annotator disagreement. The dataset contains 1\,457 facial images.
The annotation set from the original paper~\cite{joint_acne_grading_ldl} provides local lesion counts and global acne severity scores together with 18\,983 unlabeled bounding boxes. The second annotation set~\cite{acneai} covers 1\,204 images after removing low-quality images and contains 32\,443 unlabeled bounding boxes; experts were instructed to annotate more acne lesions.

\item \textbf{AcneSCU}~\cite{high_quality_proposals_acne_detection}.
We include AcneSCU to study labeled annotations in the same application domain. It contains 276 facial images acquired with the VISIA complexion analysis system and 31\,777 instance-segmentation annotations across 10 lesion categories.

\item \textbf{TexBiG}~\cite{krippendorf_obj_det}.
We include TexBiG to evaluate the approach in a substantially different domain with multi-annotator data. This historical-document layout dataset contains more than 52\,000 expert-annotated instances across 19 classes, with multiple annotators used for quality control.
  
\item \textbf{Perturbed datasets.} We additionally construct perturbed versions of these datasets using box dropping, box shifting, and label swapping. The perturbation level is controlled by a single parameter $\gamma$, whose interpretation depends on the perturbation type. For \textit{box drop}, $\gamma \in [0, 1]$ denotes the fraction of boxes to remove in each image. For \textit{box shift}, $\gamma$ scales the displacement of the box center. We shift each box proportionally to its size; for example, under Uniform noise the $x$-coordinate is perturbed as
  \begin{equation}
      x_{c2} = x_{c1} + \varepsilon_x, \quad \varepsilon_x \sim \mathcal{U}(-u_x, u_x), \quad u_x = \gamma w,
  \end{equation}
  similarly, under Normal noise:
\begin{equation}
      x_{c2} = x_{c1} + \varepsilon_x, \quad \varepsilon_x \sim \mathcal{N}(0, \sigma_x), \quad \sigma_x = \gamma w,
  \end{equation}
  
  and analogously for the $y$-coordinate. In the theoretical calculation for Eq.~\ref{eq:cdf_final}, symmetry
allows us to work with the nonnegative absolute displacements
$|\varepsilon_x|$ and $|\varepsilon_y|$. For \textit{label swap}, $\gamma$ specifies the fraction of boxes in an image whose labels are to be exchanged. Selected boxes are paired at random and their labels are swapped within each pair.
\end{itemize}

\begin{figure}[tb]
  \centering
  \begin{subfigure}{0.49\linewidth}
    \includegraphics[width=\linewidth]{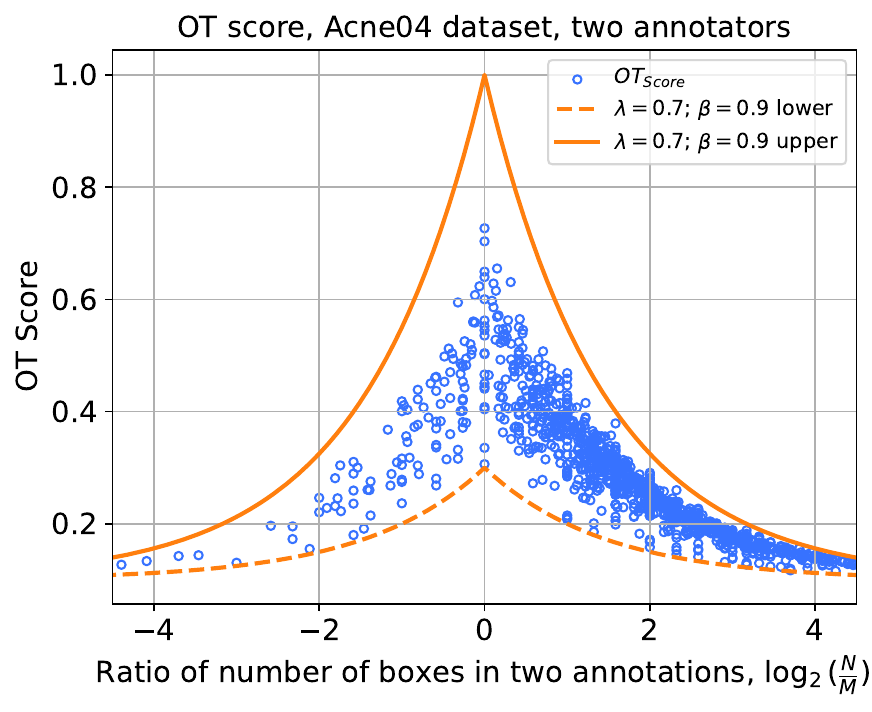}
    \caption{}
    \label{fig:acne_vs_ratio}
  \end{subfigure}
  \hfill
  \begin{subfigure}{0.49\linewidth}
    \includegraphics[width=\linewidth]{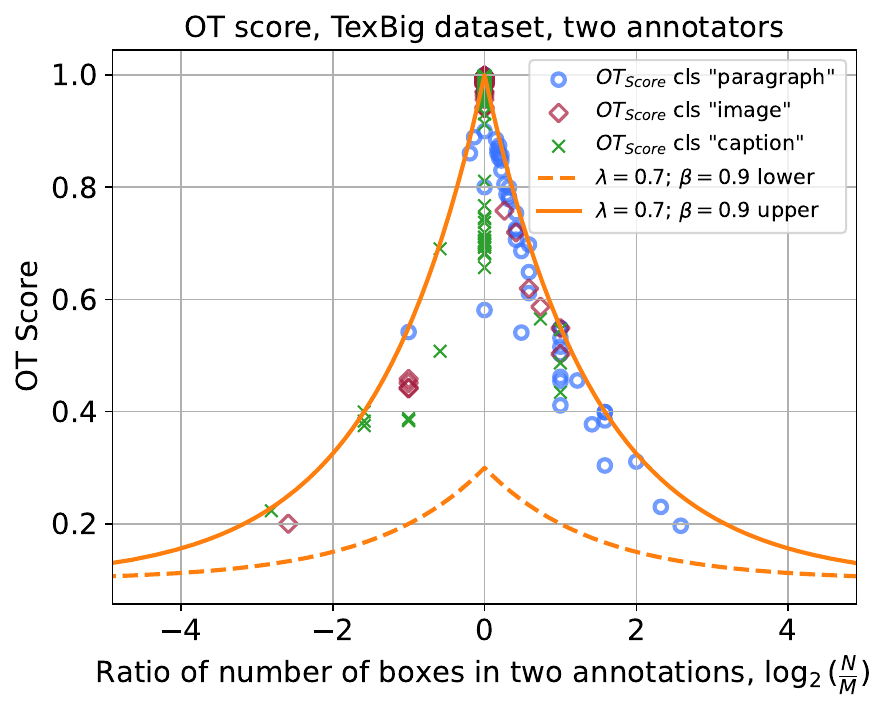}
    \caption{}
    \label{fig:texbig_vs_ratio}
  \end{subfigure}

  \vspace{0mm}
  \begin{subfigure}{0.49\linewidth}
    \includegraphics[width=\linewidth]{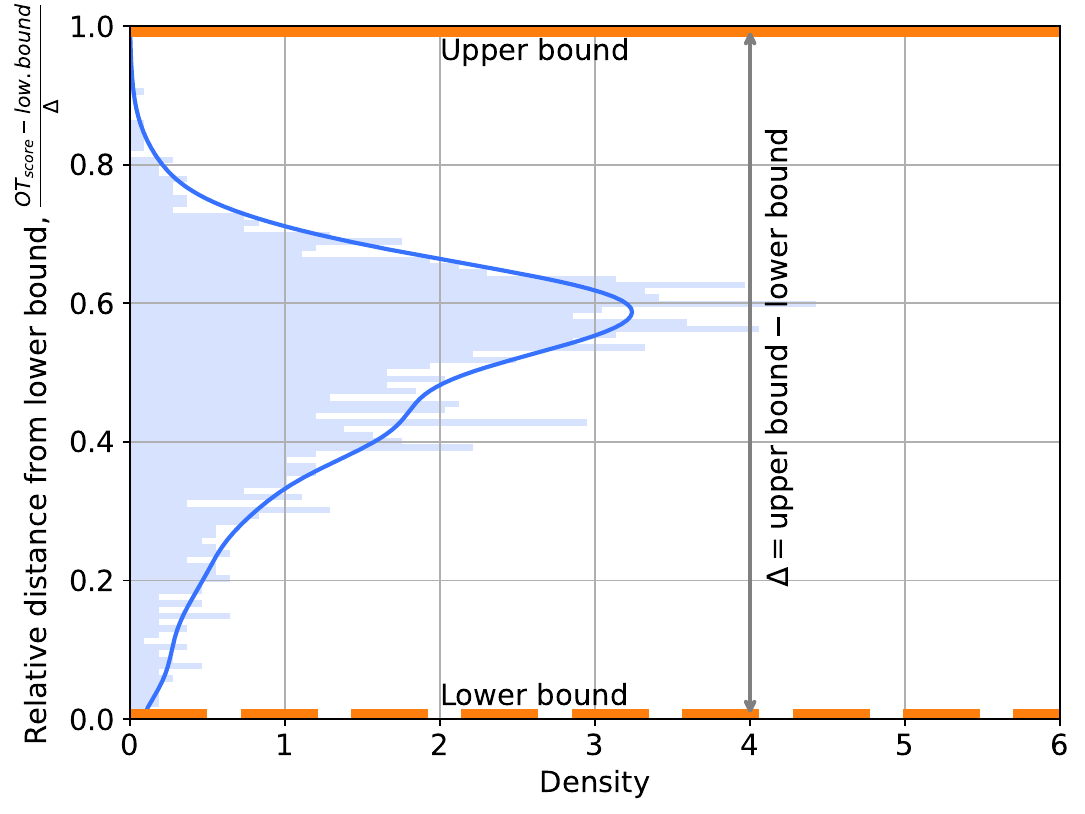}
    \caption{}
    \label{fig:label_3}
  \end{subfigure}
  \hfill
  \begin{subfigure}{0.49\linewidth}
    \includegraphics[width=\linewidth]{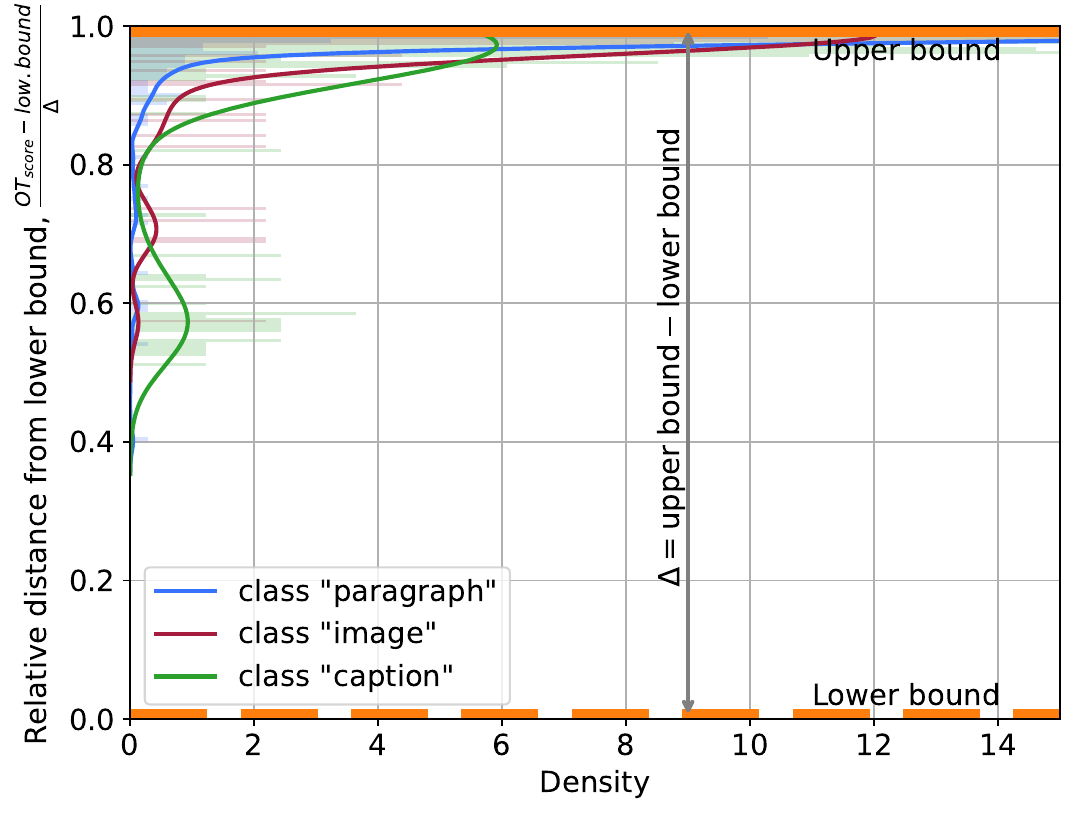}
    \caption{}
    \label{fig:label_4}
  \end{subfigure}
\caption{OT scores relative to their theoretical bounds.
\textbf{a)} ACNE04 scores span much of the feasible range.
\textbf{b)} OT scores for several TexBiG classes.
\textbf{c,d)} Score distributions between the lower and upper bounds
for ACNE04 and TexBiG, respectively. Because TexBiG objects are larger
and easier to localize, their spatial costs are generally smaller and
their scores lie closer to the upper bound.}
\label{fig:ot_vs_ratio}
\end{figure}

\subsection{Results and Discussion}
Fig.~\ref{fig:ot_vs_ratio} shows example OT scores comparing two annotations for the ACNE04 dataset (Fig.~\ref{fig:acne_vs_ratio}) and for the TexBiG dataset (Fig.~\ref{fig:texbig_vs_ratio}), along with the theoretical boundaries in Eq.~\ref{eq:boundaries}. There are two key differences between these sets of annotations. First, acne objects generally have a smaller size relative to the image resolution than objects on paper (such as paragraphs or headings), meaning that changes in localization affect the cost metric much more strongly in the former case. Second, when an object is easy to annotate (e.g.\ a clearly visible paragraph in a document), it can naturally be surrounded by a rectangular box. From the figures one can clearly see that, while for ACNE04 we obtain a wide range of OT values, for TexBiG almost all objects lie close to the upper bound: localization error rarely becomes crucial.
Based on these observations, we believe that the OT score can be applied to any dataset, but it is most useful and informative when there are many small or medium-sized boxes in specialized domains (e.g.\ acne-lesion detection), where objects can be easily missed or misclassified. 

\begin{figure}[tb]
  \centering
  \begin{subfigure}{0.49\linewidth}
    \includegraphics[width=\linewidth]{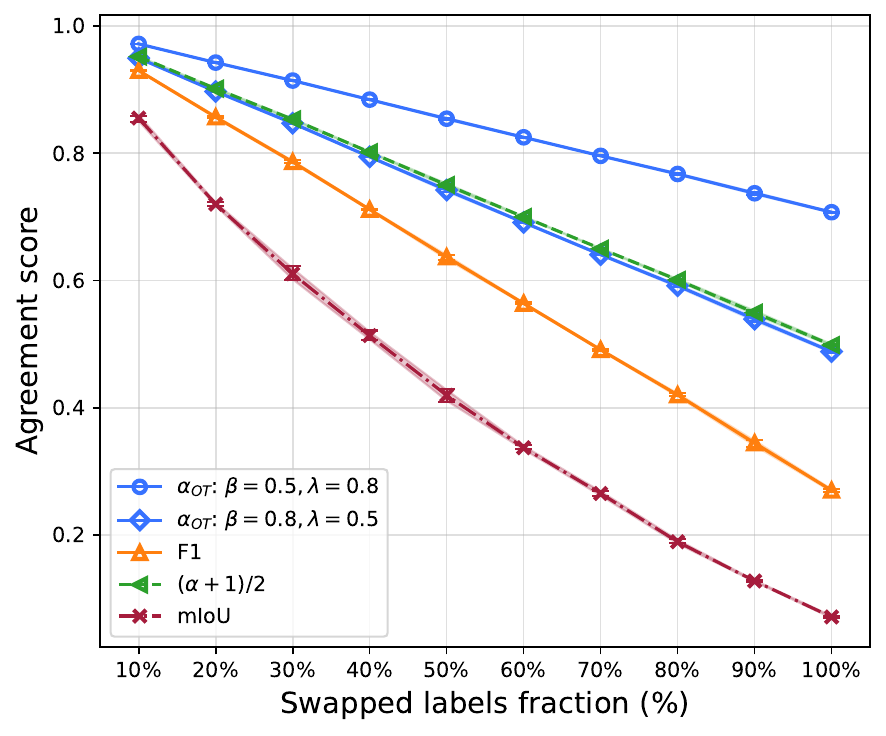}
    \caption{}
    \label{fig:vs_swap}
  \end{subfigure}
  \hfill
  \begin{subfigure}{0.49\linewidth}
    \includegraphics[width=\linewidth]{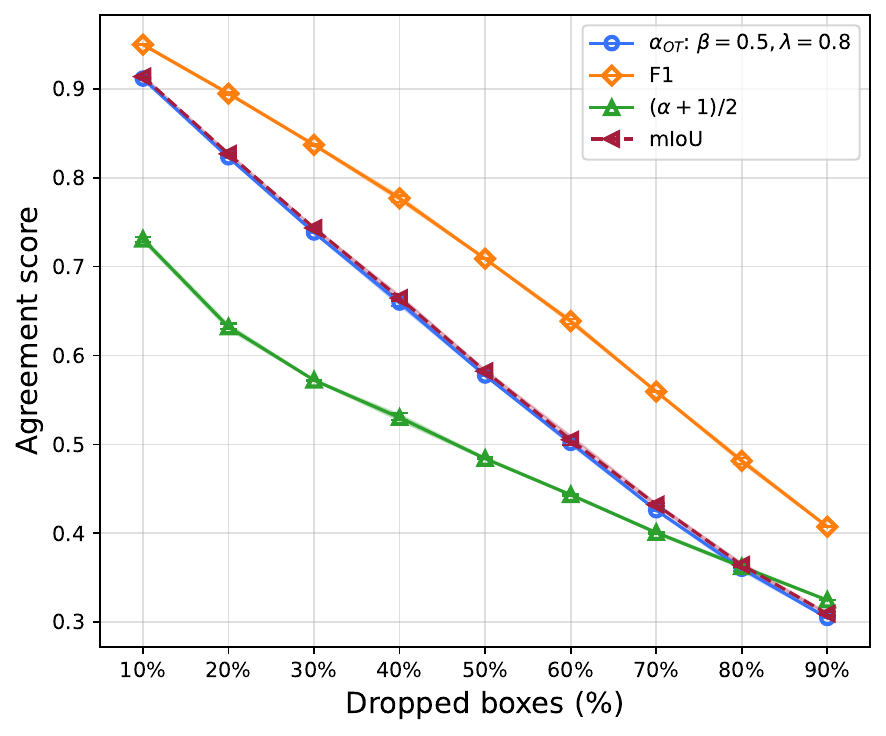}
    \caption{}
    \label{fig:vs_drop}
  \end{subfigure}

  \vspace{0pt}
  \begin{subfigure}{\linewidth}
    \centering
    \includegraphics[width=1\linewidth]{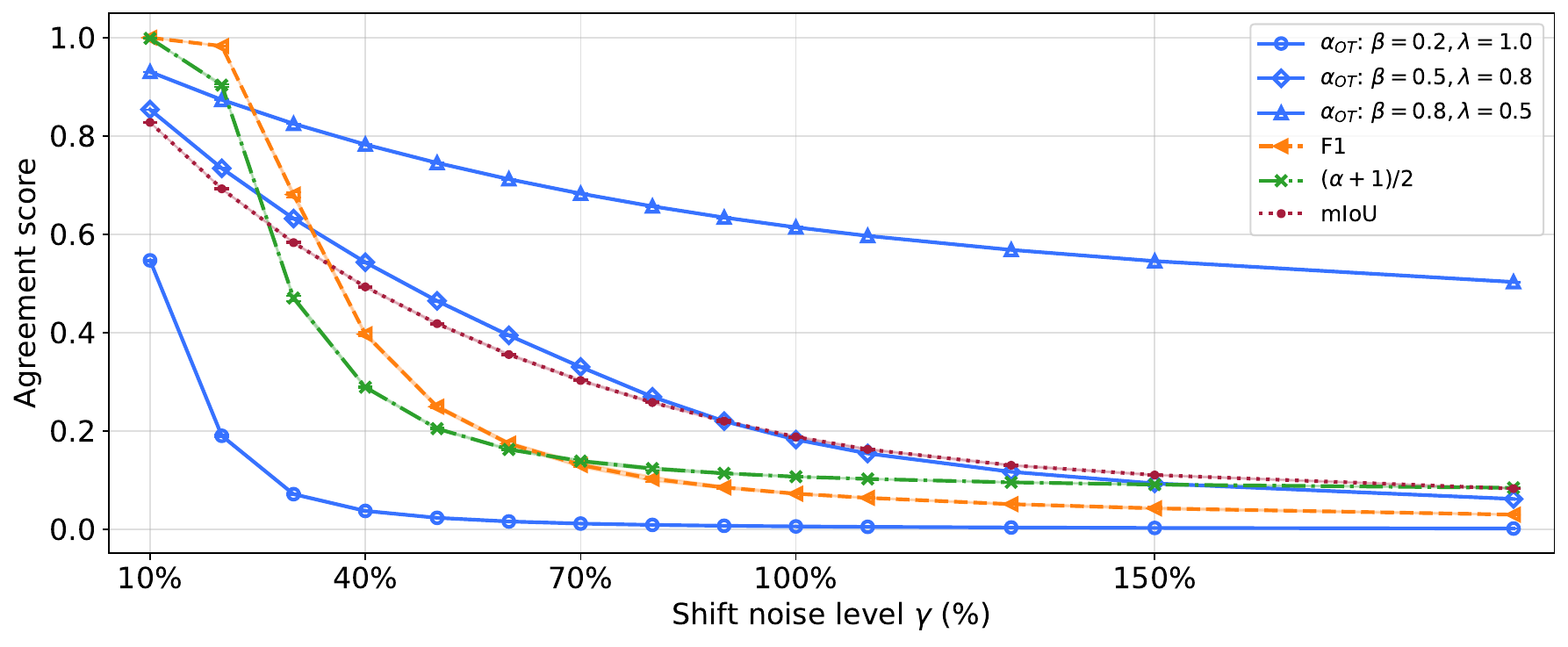}
    \caption{}
    \label{fig:vs_shift}
  \end{subfigure}

  \caption{Annotator agreement between original and perturbed annotations. \textbf{a)} Agreement versus fraction of swapped labels on AcneSCU dataset. \textbf{b)} Agreement versus ratio of dropped boxes on ACNE04 dataset. \textbf{c)} Agreement versus shift noise level on ACNE04 dataset.
  }
  \label{fig:discrepancy}
\end{figure}

Fig.~\ref{fig:discrepancy} shows the dependence of the metrics on the noise level $\gamma$. We use the method for computing annotator discrepancy/similarity from \cite{discrepancy_ratio}. If the annotation set from annotator $j$ on image $i$ is
$\mathcal{Y}_j^i$ and $A_i$ is the number of annotators for image $i$,
then for a pairwise similarity measure
$M(\mathcal{Y}_j^i,\mathcal{Y}_{j'}^i)$, such as F1,
Krippendorff's $\alpha$, or mIoU, annotator similarity is defined as
\begin{equation}
    S(\{\mathcal{Y}_j^i\})=\frac{1}{N}\sum_{i=1}^N\frac{1}{A_i(A_i-1)}\sum_{j,j'=1, j\neq j'}^{A_i}M(\mathcal{Y}_j^i, \mathcal{Y}_{j'}^i)
    \label{eq:discrep}
\end{equation}
In this experiment, we test how each metric behaves as the perturbation
level increases by comparing the original annotation with a perturbed
one. Thus, $A_i=2$ in Eq.~\ref{eq:discrep}, which reduces to the mean
pairwise similarity across images. For OT, we use the dataset-level
chance-corrected coefficient from Eq.~\ref{eq:chance_corr}. Because these metrics are not all defined on the same chance-corrected scale, we compare their response to increasing perturbation rather than their absolute values.

Fig.~\ref{fig:vs_shift} shows that IoU-threshold-based metrics exhibit
a rapid drop as noise increases, particularly once box overlaps fall
below the IoU threshold required for a true-positive match. In contrast,
the $\alpha_{\rm OT}$ curves with $\beta=0.5$ and $0.8$
decrease more gradually, reflecting their sensitivity to spatial
displacement and the absence of hard matching thresholds. The
$\beta=0.2$ setting also exhibits a rapid decrease because the smaller
dummy-matching penalty makes real-box matches less preferable even under
modest perturbations. For comparison, we also report mean IoU (mIoU),
which decreases more smoothly than threshold-based metrics but becomes
insensitive once boxes no longer overlap and is not commonly used as an
object-detection agreement metric.

Figures~\ref{fig:vs_drop} and~\ref{fig:vs_swap} illustrate the behavior
of the metrics under box dropping (ACNE04) and label swapping (AcneSCU),
respectively. In both cases, the curves are approximately linear. For
the OT-based metric, this is because each unmatched box contributes a
penalty of $\beta$ in the \textit{box drop} setting, while in the
\textit{label swap} setting a discrete label cost contributes to the
pairwise cost in Eq.~\ref{eq:ldb_dist}; thus, the total cost scales
approximately linearly with the number of such errors. An interesting
extension would be to allow graded label distances or similarities
rather than values restricted to $\{0,1\}$, and to account for label
imbalance, as suggested, for example, in~\cite{boriah2008similarity}.
We report several values of $\beta$; in practice, $\beta$ can be selected
empirically or calibrated using approaches such as
K$\alpha$LOS~\cite{tschirschwitz2026kalos}
(see supplementary material). We repeated all experiments across several
random seeds; variability was small and is omitted for clarity.
\begin{figure}[tb]
  \centering
  \begin{subfigure}{\linewidth}
    \centering
    \includegraphics[width=1\linewidth]{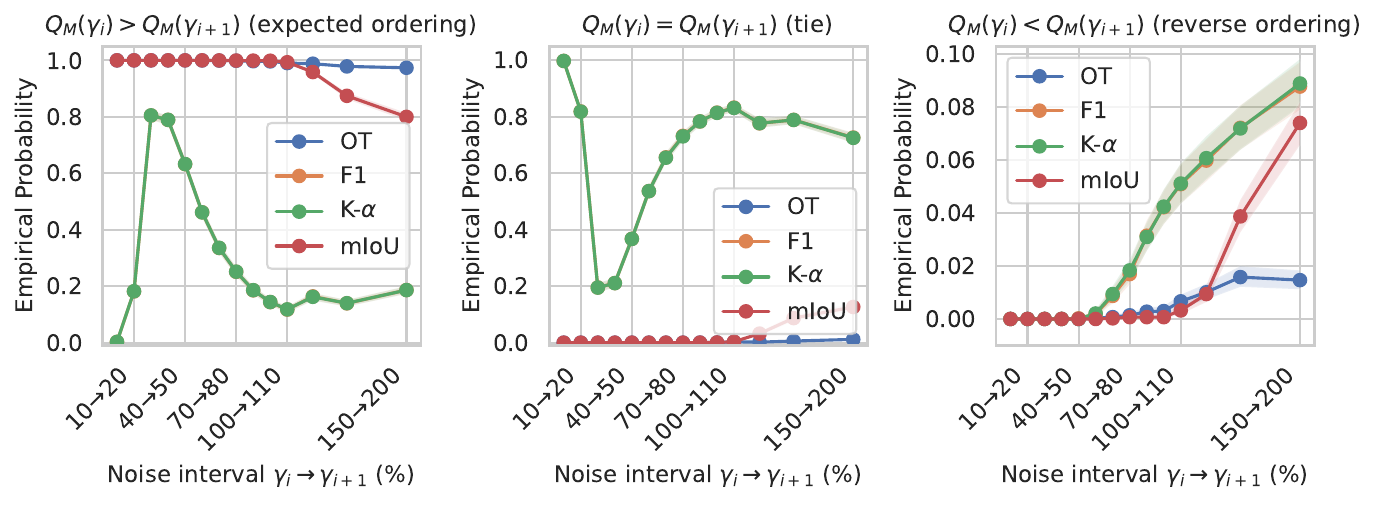}
    \caption{}
    \label{fig:ordering}
  \end{subfigure}
  
  \vspace{0pt}
  
  \begin{subfigure}{0.45\linewidth}
    \includegraphics[width=\linewidth]{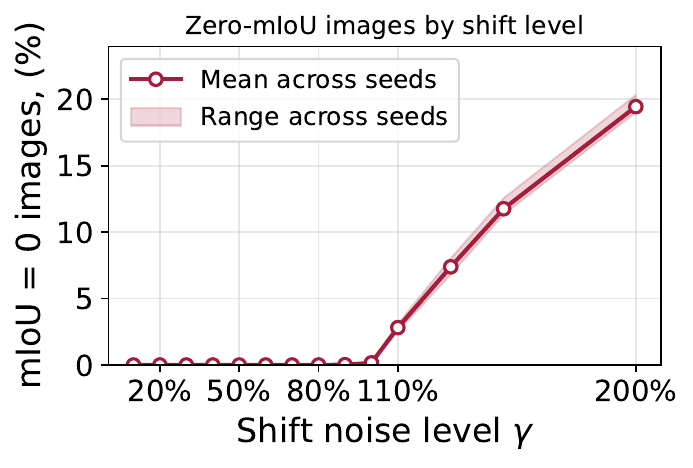}
    \caption{}
    \label{fig:miou_count}
  \end{subfigure}
  \hfill
  \begin{subfigure}{0.45\linewidth}
    \includegraphics[width=\linewidth]{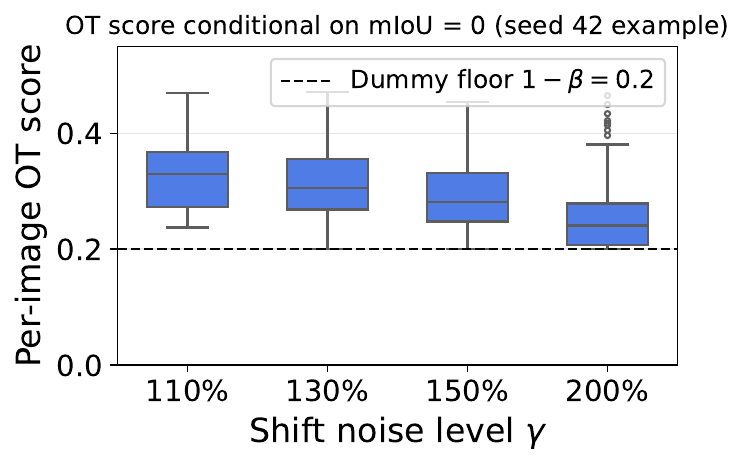}
    \caption{}
    \label{fig:miou_distr}
  \end{subfigure}
    \caption{Behavior of agreement metrics under increasing box-shift noise and beyond the $mIoU=0$ regime. \textbf{a)} Ordering under increasing shift noise. \textbf{b)} Percentage of images with $mIoU=0$ versus noise level. \textbf{c)} Distribution of OT scores for cases with $mIoU=0$.}
  \label{fig:collapse_miou}
\end{figure}

We quantify how consistently each metric responds to increasing noise.
Per-image Kendall's $\tau_b$ between increasing noise and decreasing
agreement is $0.998$ for OT and $0.993$ for mIoU, compared with $0.790$
for F1 and $0.790$ for Krippendorff's $\alpha$.

We additionally examine ordering between adjacent noise levels at the
dataset level. Let $Q_M(\gamma)$ denote the dataset-level value for
metric $M$ at noise level $\gamma$. For pairwise similarity measures,
$Q_M$ is the mean similarity in Eq.~\ref{eq:discrep}; for OT, it is
$\alpha_{\rm OT}$ from Eq.~\ref{eq:chance_corr}. For adjacent noise
levels $\gamma_i<\gamma_{i+1}$, define
$\Delta_M=Q_M(\gamma_i)-Q_M(\gamma_{i+1})$. We classify each comparison
as expected ordering if $\Delta_M>\varepsilon$, a tie if
$|\Delta_M|\leq\varepsilon$, and reverse ordering if
$\Delta_M<-\varepsilon$, with $\varepsilon=10^{-3}$.
Fig.~\ref{fig:ordering} shows that OT and mIoU preserve the expected
ordering in most cases, whereas the threshold-based metrics exhibit many
ties: ceiling saturation dominates at low noise and floor saturation at
high noise. Reversals at the largest shifts can occur when perturbed
boxes move closer to neighboring reference boxes and are rematched.

Both mIoU and OT provide graded responses to localization changes
while boxes overlap. Once overlap vanishes, however, mIoU is identically
zero regardless of further displacement, whereas OT can retain graded
spatial information when $\beta/\lambda>1/2$. On ACNE04,
Fig.~\ref{fig:miou_count} shows that the fraction of images with
$mIoU=0$ rises to about 19\% at 200\% shift. For these images, OT scores
span approximately $0.20$--$0.47$ for $\beta=0.8$, as shown in
Fig.~\ref{fig:miou_distr}. Thus OT retains spatial resolution even after
mIoU reaches zero.

These synthetic perturbations are controlled stress tests rather than models of human annotation noise. ACNE04 and TexBiG provide real annotation disagreements, but we do not have independent expert ratings of disagreement severity; such validation remains future work. Moreover, $\lambda$ and $\beta$ are task-dependent. Supplementary Fig.~3 illustrates their effect on spatial matching, while Supplementary Fig.~4 shows data-driven calibration of $\beta$ on TexBiG.

\section{Conclusion}
\label{sec:conclusion}
In this work, we adapted an Optimal Transport (OT)–based metric for quantifying annotator agreement in object detection. By formulating agreement as a transport problem between sets of bounding boxes, our approach naturally integrates spatial and label information and allows for flexible handling of missing or extra annotations.

We analyzed several theoretical properties of the OT-based metric, deriving bounds for cases in which the transport plan is known and characterizing the possible range of OT scores under specific parameter settings. We also studied the behavior of the metric as a function of $\beta$ under a noisy-box assumption. This analysis helps to understand how many real matches remain as $\beta$ or the noise level increases, how to select $\beta$ for a given task, and when deviations from the theoretical assumptions signal that annotations have become substantially different.

Through experiments on both real-world and synthetically perturbed datasets, we demonstrated that the OT-based metric is particularly sensitive to differences in annotations, especially with respect to the spatial positions of bounding boxes. In scenarios where traditional IoU threshold-based metrics (such as F1 or Krippendorff’s $\alpha$) fail to distinguish between qualitatively different annotation patterns, OT still reflects meaningful variation in annotator behavior.

Our findings lead to two main conclusions. First, model performance evaluation and annotator agreement assessment are closely related but not identical: what is treated as an “error’’ in model evaluation does not necessarily correspond to disagreement between annotators, and OT can help to disentangle these notions. Second, because the main sensitivity of the OT cost comes from spatial localization, we believe it is especially well suited to domains with many small or medium-sized boxes and non-trivial domain knowledge (e.g. medical imaging), where objects are easily missed or misclassified.

Overall, these results indicate that the OT metric provides a richer and more informative measure of annotator agreement and discrepancy, and can support more reliable model validation in the presence of noisy or conflicting labels.

\bibliographystyle{splncs04}
\bibliography{main}

\end{document}


\title{Supplementary Material:\\Quantifying Annotation Agreement in Object Detection with Optimal Transport}

\titlerunning{Optimal Transport for Annotation Agreement}

\author{Kirill Prokhorov\inst{1}\orcidlink{0000-0001-5915-6614} \and
Alexandr A. Kalinin\inst{2}\orcidlink{0000-0003-4563-3226}}

\authorrunning{K.~Prokhorov et al.}

\institute{Openface.io, Vilnius, Lithuania\\
\email{prokhorov@openface.me}
\and
Broad Institute of MIT and Harvard, Cambridge, MA, USA}

\maketitle

\section{On the evaluation of $P(c_{ij}<\beta)$}
To calculate the probability that the cost $c_{ij}$ is less than $\beta$, we evaluate
\begin{equation}
  P(c_{ij}<\beta) = F_c(\beta) = \int_{0}^{\beta} f_c(t) \, dt = 
  \iint_{c(\xi, \eta) < \beta} f_{|\varepsilon_x|, |\varepsilon_y|}(\xi, \eta) \, d\xi d\eta,
  \label{eq:prob}
\end{equation}
where $f_c$ is the PDF of the cost and $f_{|\varepsilon_x|, |\varepsilon_y|}$ is the joint PDF of the absolute noise components.

\begin{figure}[tb]
  \centering
  \includegraphics[width=0.8\linewidth]{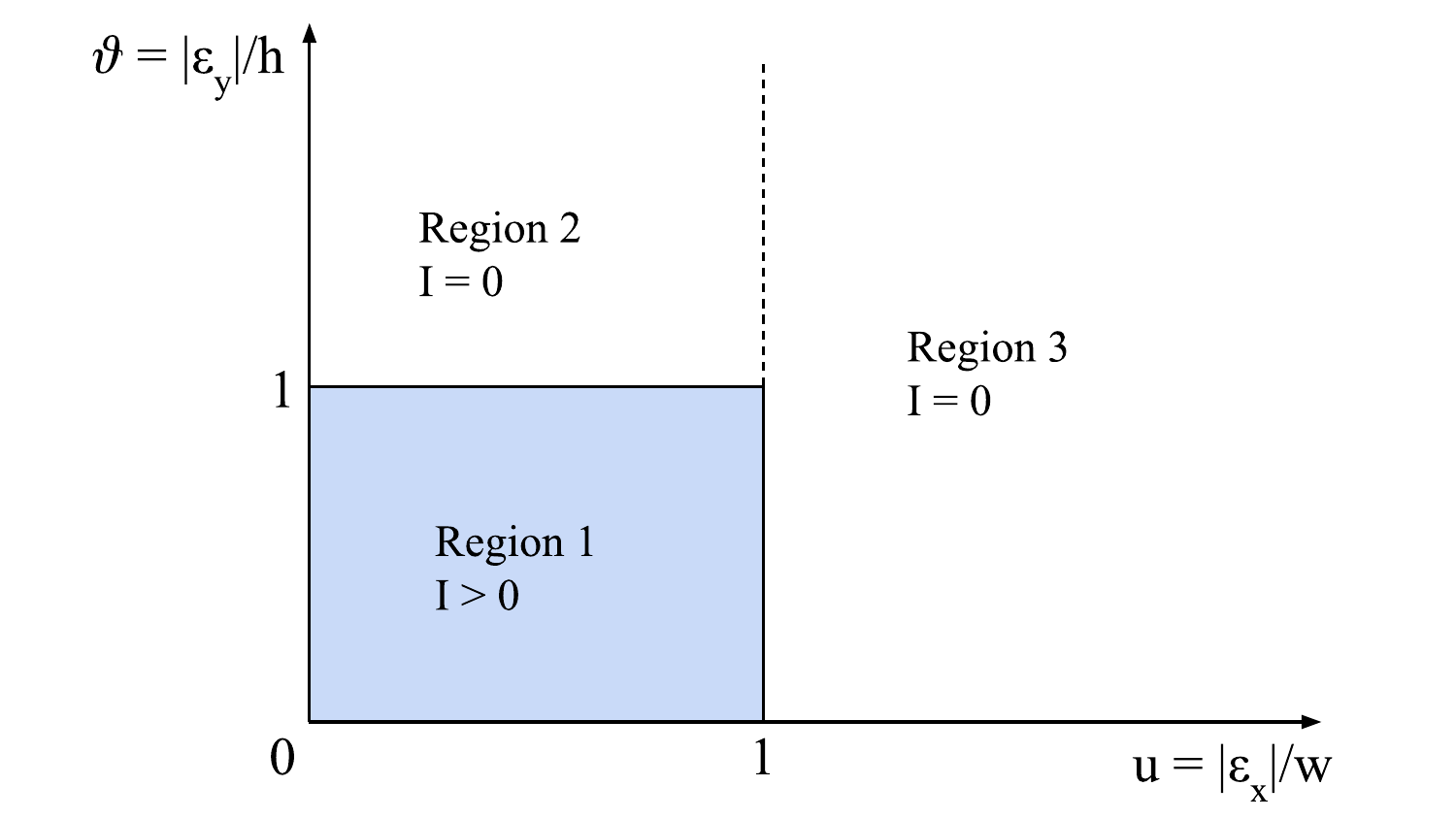}
  \caption{Integration regions used in the calculation in Eq.~\ref{eq:prob}.}
  \label{fig:regions}
\end{figure}

To characterize the integration domain $c(\xi, \eta) < \beta$, we use the expression for GIoU in the cases where the two boxes do and do not intersect. This leads to a decomposition of the integral in Eq.~\ref{eq:prob} into three regions (Fig.~\ref{fig:regions}): $u<1,\, \vartheta<1$, where $I>0$, and two regions with $I=0$, namely $u<1,\, \vartheta\geq1$ and $u\geq1,\, \vartheta>0$. For $I=0$ we have:

\begin{equation}
  GIoU_{I=0} = -\frac{(1+u)(1+\vartheta) - 2}{(1+u)(1+\vartheta)}= 1-2\beta,
\end{equation} 
which gives
\begin{equation}
  \vartheta_{I=0}(u) = \frac{1}{(1-\beta)(1+u)}-1.
\end{equation}
The same can be done for the case $I\neq 0$:

\begin{equation}
  GIoU_{I\neq 0} =
  \frac{(1-u)(1-\vartheta)}{2-(1-u)(1-\vartheta)} -
  \frac{(1+u)(1+\vartheta)-2+(1-u)(1-\vartheta)}
       {(1+u)(1+\vartheta)}
  = 1-2\beta.
  \label{eq:giou_nonzero}
\end{equation}

For $I\neq0$, Eq.~\ref{eq:giou_nonzero} can be rewritten as a quadratic
equation in $\vartheta$,
\begin{equation}
A(u,\beta)\vartheta^2+B(u,\beta)\vartheta+C(u,\beta)=0,
\end{equation}
where
\begin{align}
A(u,\beta) &= (u-1)\bigl[\beta(u+1)-2u-1\bigr],\\
B(u,\beta) &= (u+1)(u+1-2\beta),\\
C(u,\beta) &= -(u+1)\bigl[\beta(u+1)-u\bigr].
\end{align}
We denote the relevant (larger) real root by
$\vartheta_{I\neq0}(u)$. When defining the integration limit below,
we clip this root to the feasible interval $[0,1]$.

We can now define the integration limits in each region. For
$u<1,\,\vartheta<1$, we set
$a_1(u)=\max\!\left(0,\min\!\left(1,\vartheta_{I\neq0}(u)\right)\right)$;
for $u<1,\,\vartheta\geq1$, we set
$a_2(u)=\max\!\left(1,\vartheta_{I=0}(u)\right)$; and for
$u\geq1,\,\vartheta>0$, we set
$a_3(u)=\max\!\left(0,\vartheta_{I=0}(u)\right)$.

Substituting these region-specific limits into Eq.~\ref{eq:prob}
decomposes the integration domain into the three regions described
above:
\begin{multline}
  F_c(\beta) =
  \int_{0}^{1}\int_{0}^{a_1(\xi)}
  f_{u,\vartheta}(\xi,\eta)\,d\eta\,d\xi +
  \int_{0}^{1}\int_{1}^{a_2(\xi)}
  f_{u,\vartheta}(\xi,\eta)\,d\eta\,d\xi + \\
  + \int_{1}^{\infty}\int_{0}^{a_3(\xi)}
  f_{u,\vartheta}(\xi,\eta)\,d\eta\,d\xi .
\end{multline}

If the noise components are independent, the joint distribution
factorizes as
\begin{equation}
  f_{u,\vartheta}(\xi,\eta)
  = f_u(\xi)\,f_{\vartheta}(\eta),
\end{equation}
so the inner integrals can be expressed using the CDF
$F_{\vartheta}$:
\begin{multline}
  F_c(\beta) =
  \int_{0}^{1} f_u(\xi)\,
  F_{\vartheta}\bigl(a_1(\xi)\bigr)\,d\xi +
  \int_{0}^{1} f_u(\xi)\,
  \bigl[F_{\vartheta}\bigl(a_2(\xi)\bigr)
  - F_{\vartheta}(1)\bigr]\,d\xi + \\
  + \int_{1}^{\infty} f_u(\xi)\,
  F_{\vartheta}\bigl(a_3(\xi)\bigr)\,d\xi .
  \label{eq:cdf_final}
\end{multline}

\section{Examples of spatial properties of the OT score}
\begin{figure}[tb]
  \centering
  \begin{subfigure}{0.48\linewidth}
    \includegraphics[width=\linewidth]{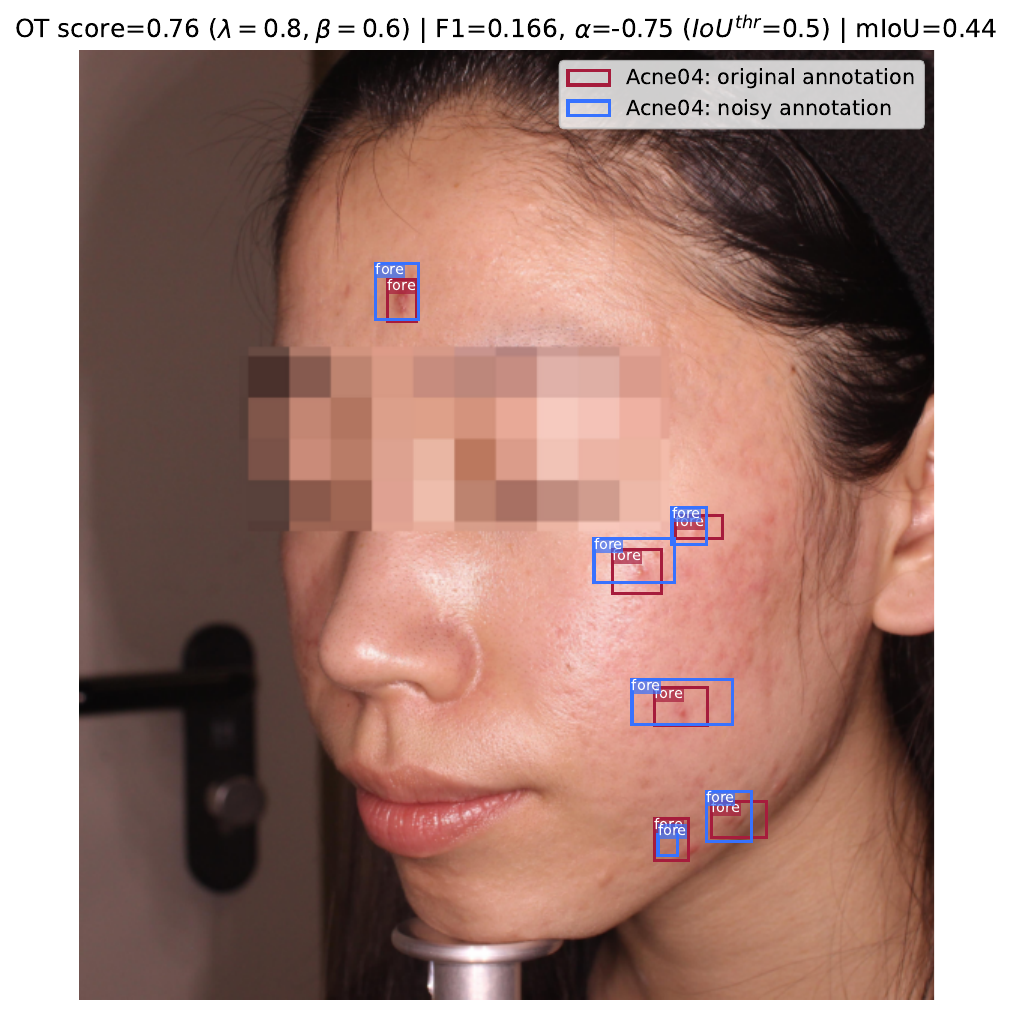}
    \caption{}
    \label{fig:shifted_boxes_1}
  \end{subfigure}
  \hfill
  \begin{subfigure}{0.48\linewidth}
    \includegraphics[width=\linewidth]{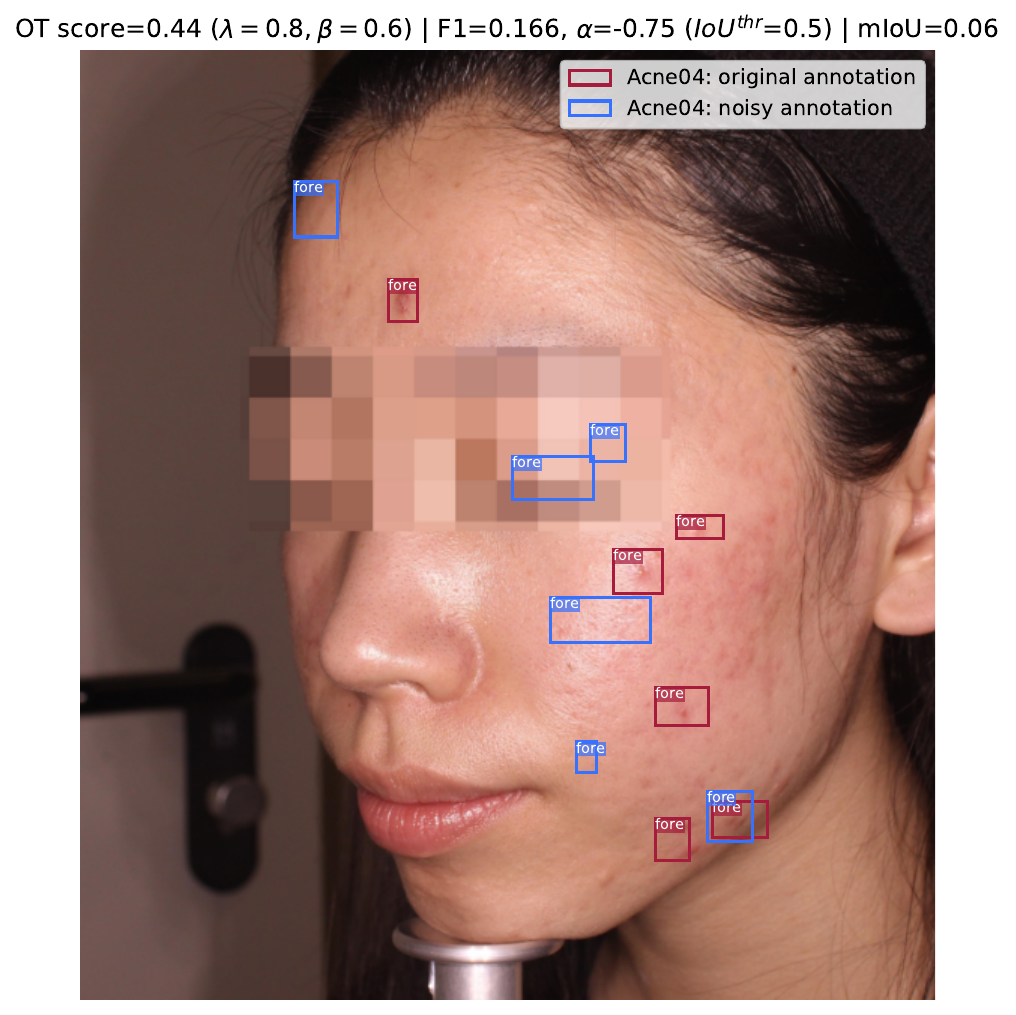}
    \caption{}
    \label{fig:shifted_boxes_2}
  \end{subfigure}
  \caption{\textbf{a)} Two annotations with a moderate spatial shift.
\textbf{b)} One annotation is deliberately shifted so that only one
pair of boxes overlaps. F1 and Krippendorff's $\alpha$ assign the same
scores to both examples, whereas OT distinguishes their spatial
disagreement. mIoU remains sensitive only while boxes overlap.}
  \label{fig:shifted_boxes}
\end{figure}

Fig.~\ref{fig:shifted_boxes} shows two annotation pairs with the same threshold-based scores but different spatial structure. In Fig.~\ref{fig:shifted_boxes_1}, the annotations differ by a moderate spatial shift: the objects are localized approximately correctly, although the boxes are not perfectly aligned. In Fig.~\ref{fig:shifted_boxes_2}, one annotation has been deliberately shifted so that only one pair of boxes overlaps, while the remaining boxes are far from their corresponding objects.

F1 and Krippendorff's $\alpha$ assign the same values to both cases, because after thresholded matching there is one true positive and the remaining boxes are treated as unmatched errors. This is appropriate for counting detection errors, but it loses information that matters for annotation agreement. The first pair reflects approximate agreement with localization noise, whereas the second reflects a stronger spatial disagreement. The OT score distinguishes these cases because unmatched and poorly matched boxes still contribute through their spatial costs. Mean IoU also reflects part of this difference, but only while corresponding boxes overlap; once overlap is zero, it cannot distinguish larger spatial displacements. Thus, OT provides a graded comparison of bounding-box annotations beyond the binary matched/unmatched decisions used by thresholded metrics.

\begin{figure}[tb]
  \centering
  \includegraphics[width=0.5\linewidth]{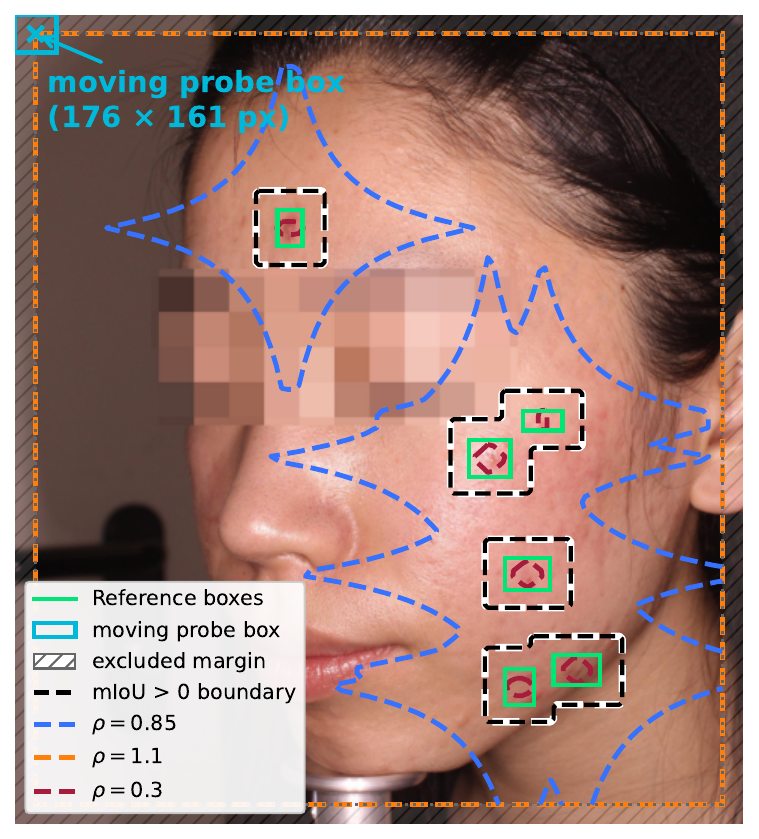}
  \caption{Spatial regions in which a translated probe box is eligible for real--real matching under different $\rho=\beta/\lambda$, compared with the region where $\mathrm{mIoU}>0$.}
  \label{fig:face_regions}
\end{figure}

In the label-free case, real matching is preferred when
$c_{ij}=\lambda(1-\mathrm{GIoU})/2<\beta$, so
$\rho=\beta/\lambda$ controls the real-to-dummy decision.
Fig.~\ref{fig:face_regions} compares the OT possible match region of a moving
probe box with the region where $\mathrm{mIoU}>0$. For $\rho\leq1/2$,
the OT region lies inside the overlap region or even reference boxes; for
$1/2<\rho<1$, it extends beyond overlap and distinguishes cases with
$\mathrm{mIoU}=0$. If $\rho>1$, all real edges are eligible, producing
$\min(M,N)$ real matches.

\section{Data-driven estimation of $\beta$ using K$\alpha$LOS}

\begin{figure}[tb]
  \centering
  \includegraphics[width=1\linewidth]{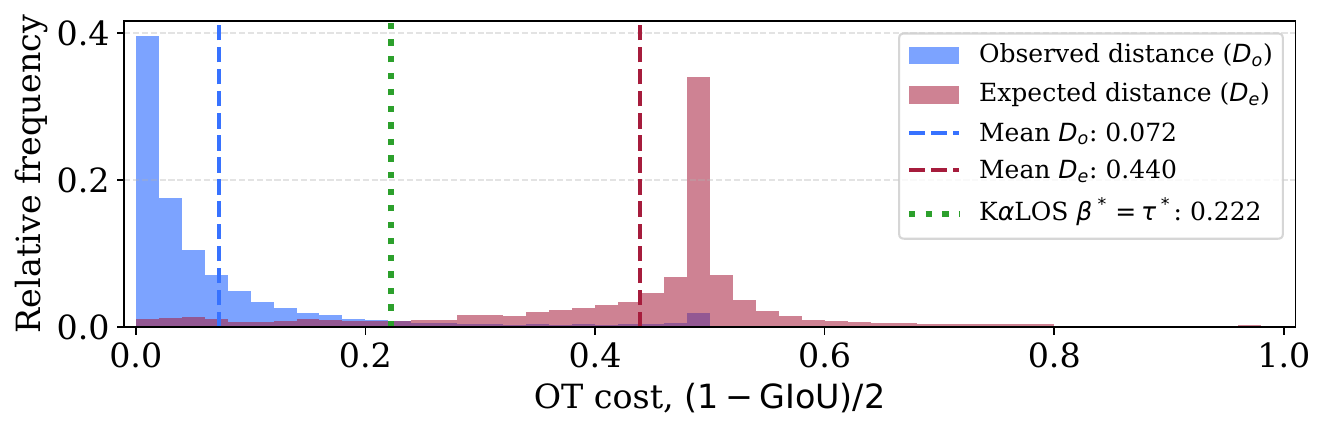}
  \caption{TexBiG observed versus expected distances.}
  \label{fig:kalos}
\end{figure}

We apply the K$\alpha$LOS data-driven $D_o/D_e$ calibration
principle~\cite{tschirschwitz2026kalos}
to real TexBiG annotations. Using all six annotators and 2,250 images
with at least two non-empty annotation sets, same-image and cross-image
\emph{box-level} spatial distances $d=(1-\mathrm{GIoU})/2$ yield
$\tau^*=0.222$, providing a data-derived candidate
$\beta^*=0.222$ for $\lambda=1$ (Fig.~\ref{fig:kalos}).
The distributions are well separated (mean $D_o=0.072$ vs.\
$D_e=0.440$), supporting calibration from real multi-annotator data.
Synthetic perturbations serve as controlled stress tests rather than
models of human annotation noise; expert severity judgments would
provide complementary validation.

\section{Additional information about datasets}
\begin{figure}[ht]
  \centering
  \includegraphics[height=6.5cm]{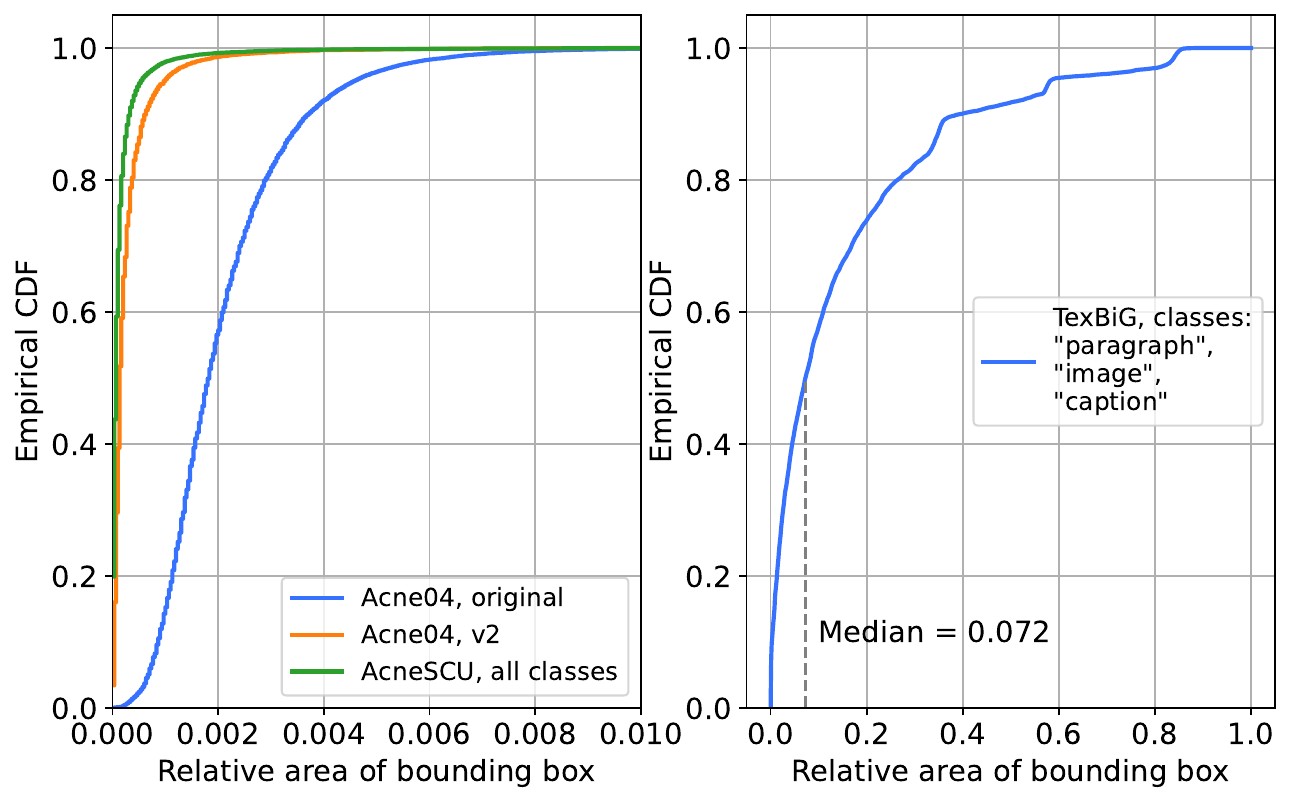}
  \caption{Empirical CDF of bounding-box area relative to image area
across the datasets and classes used in the experiments.}
  \label{fig:rel_area}
\end{figure}

Fig.~\ref{fig:rel_area} shows the empirical CDF of bounding-box area relative to image size, highlighting differences in object size across the datasets used in this paper. For Acne04 and AcneSCU, objects are relatively small: the vast majority of boxes occupy less than 1\% of the image area. In contrast, TexBiG (and in particular the three classes considered in our experiments) contains substantially larger boxes. Although many small boxes are also present, the median relative area is approximately 7\%, and there is a substantial number of objects that cover 30\% or more of the image. These differences in object size across datasets help illustrate how the proposed approach behaves in distinct, realistic scenarios.

\begin{figure}[!ht]
  \centering

  \begin{subfigure}{0.25\linewidth}
    \centering
    \includegraphics[width=\linewidth]{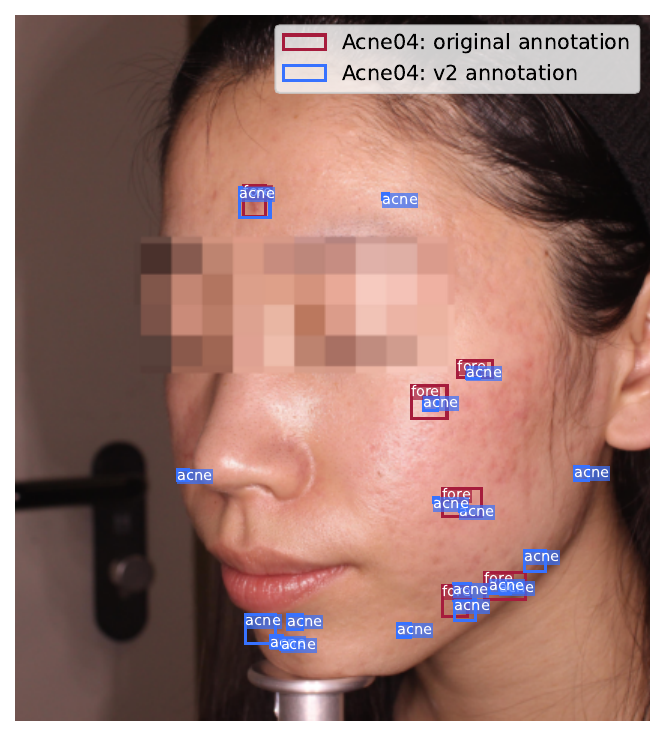}\\[-1mm]
    \textbf{(a)}
  \end{subfigure}
  \hspace{0pt}
  \begin{subfigure}{0.25\linewidth}
    \centering
    \includegraphics[width=\linewidth]{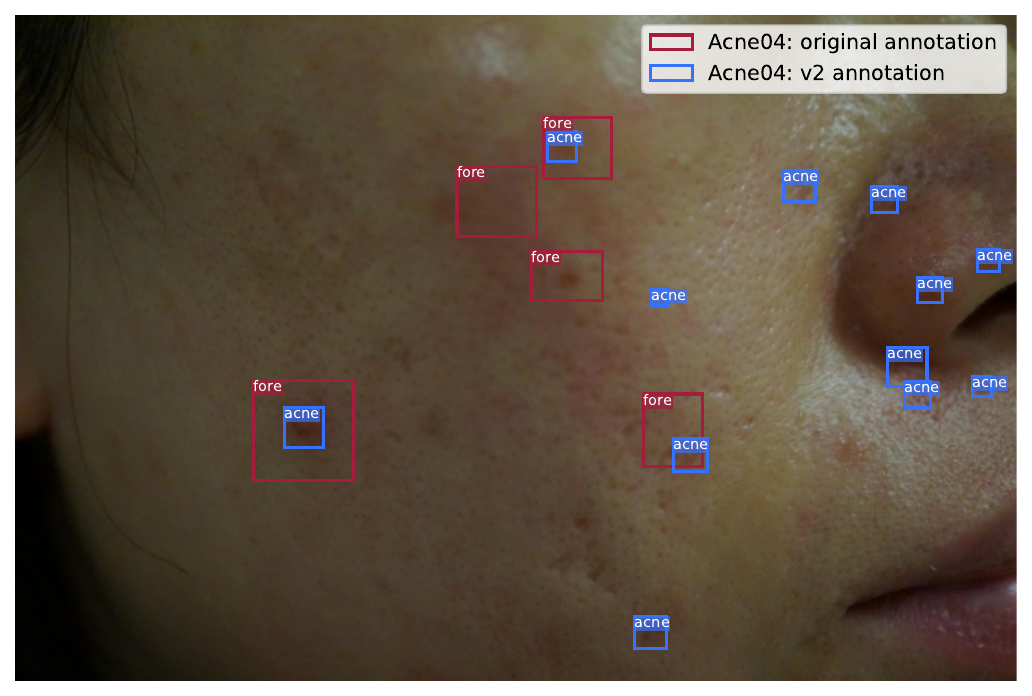}\\[-1mm]
    \textbf{(b)}
  \end{subfigure}

  \vspace{-1mm}

  \begin{subfigure}{0.25\linewidth}
    \centering
    \includegraphics[width=\linewidth]{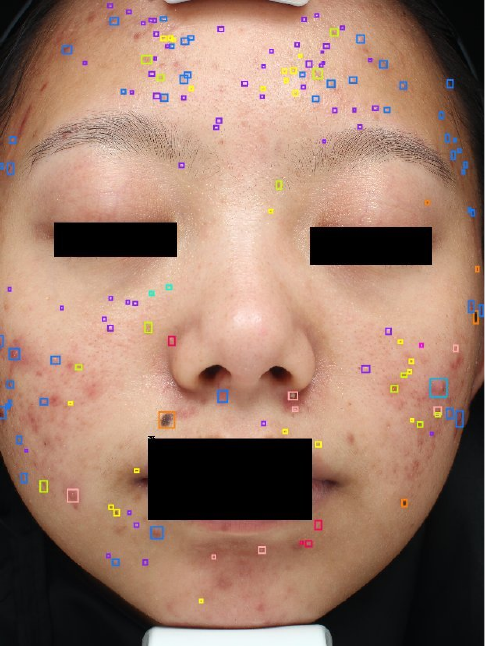}\\[-1mm]
    \textbf{(c)}
  \end{subfigure}
  \hspace{0pt}
  \begin{subfigure}{0.25\linewidth}
    \centering
    \includegraphics[width=\linewidth]{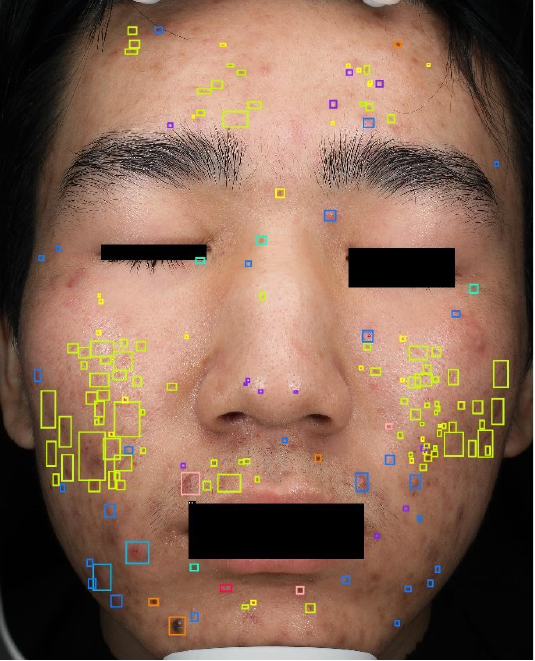}\\[-1mm]
    \textbf{(d)}
  \end{subfigure}

  \vspace{-1mm}

  \begin{subfigure}{0.4\linewidth}
    \centering
    \includegraphics[width=\linewidth]{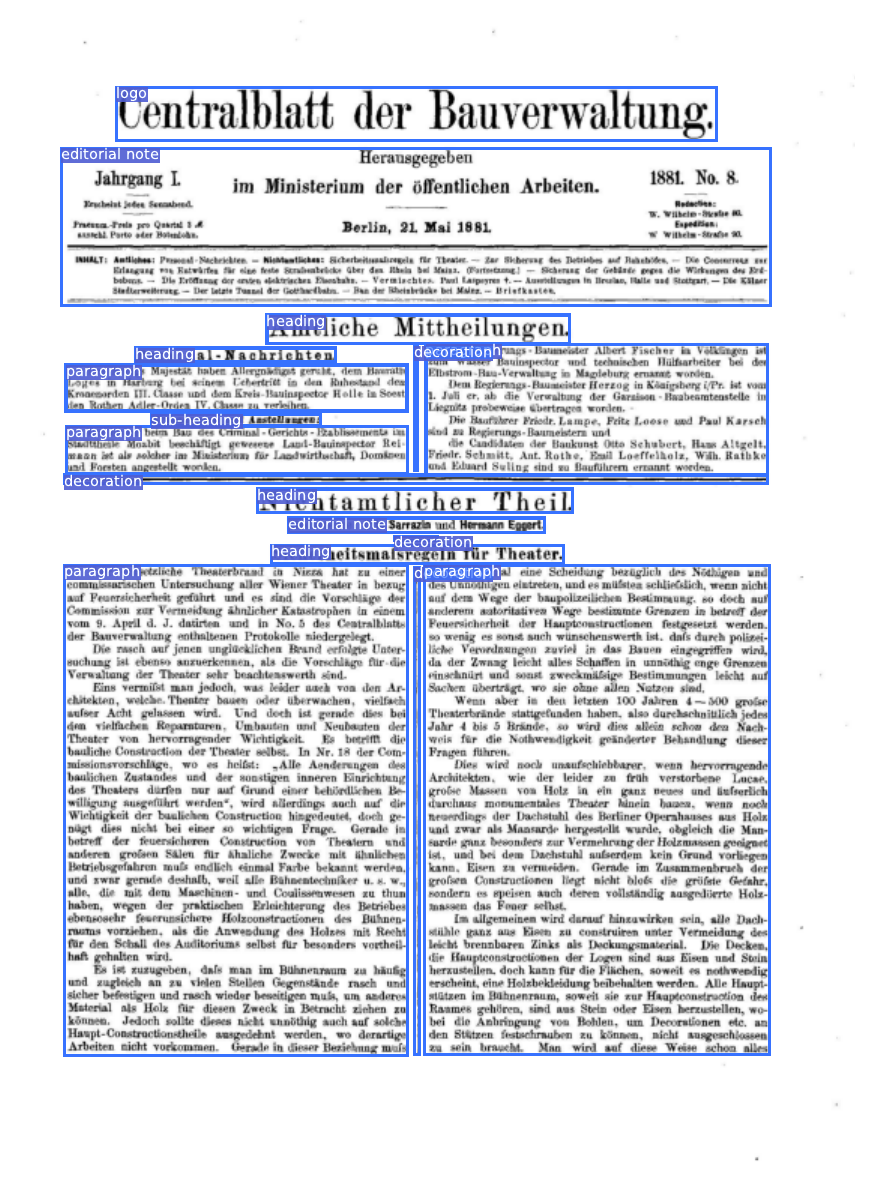}\\[-1mm]
    \textbf{(e)}
  \end{subfigure}
  \hspace{0pt}
  \begin{subfigure}{0.4\linewidth}
    \centering
    \includegraphics[width=\linewidth]{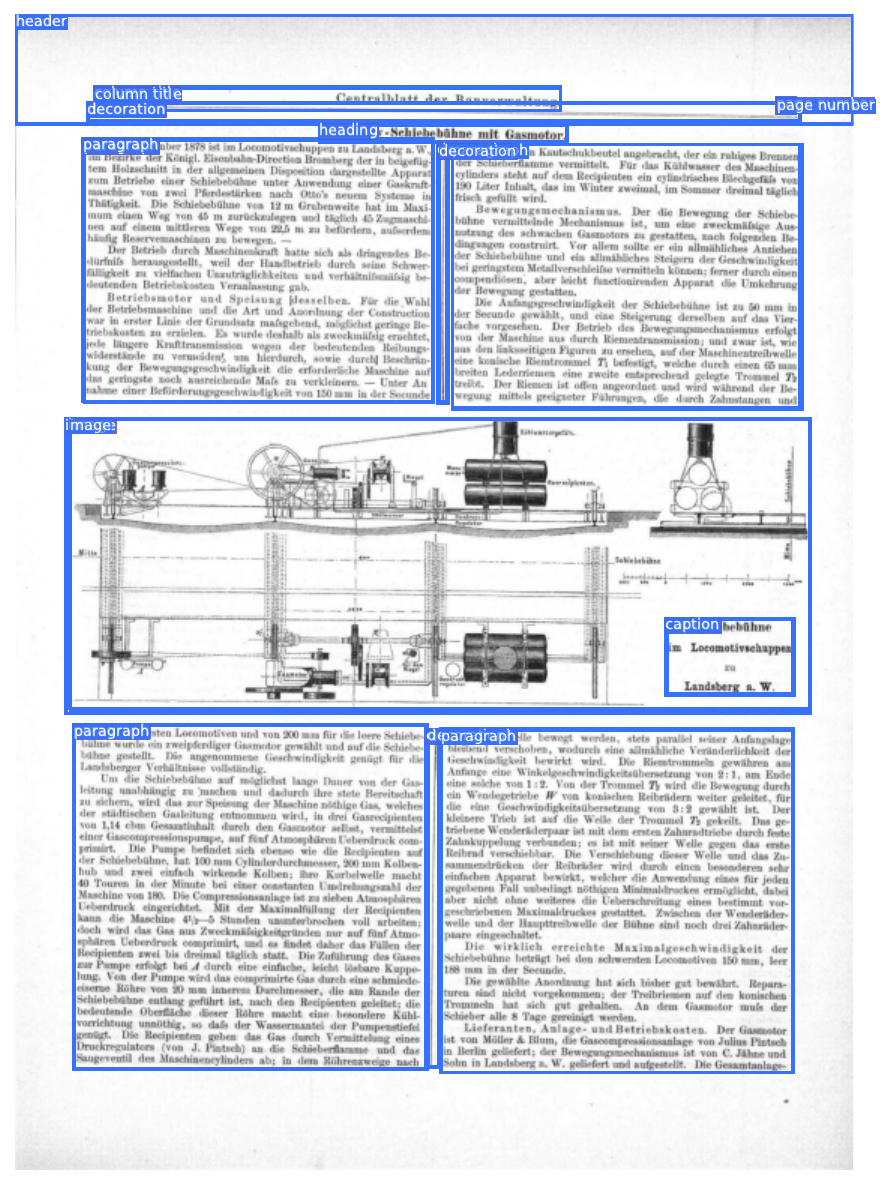}\\[-1mm]
    \textbf{(f)}
  \end{subfigure}

  \vspace{-1mm}
  \caption{Example images from the datasets.
  \textbf{a,b)} ACNE04 with two annotation sets.
  \textbf{c,d)} AcneSCU; colors denote lesion classes.
  \textbf{e,f)} TexBiG.}
  \label{fig:examples}
\end{figure}

Fig.~\ref{fig:examples} presents several example images from the datasets. \textbf{ACNE04} is an unlabeled dataset for which we used two different annotation sets from two papers. \textbf{AcneSCU} includes 10 lesion categories: open comedone, closed comedone, papule, pustule, nodule, atrophic scar, hypertrophic scar, melasma, nevus, and other. \textbf{TexBiG} contains 19 document layout classes: paragraph, decoration, page number, column title, image, header, noise, caption, heading, editorial note, frame, footer, sub-heading, equation, author, footnote, table, logo, and advertisement.
The three classes used in the main experiments (\textit{paragraph}, \textit{image}, and \textit{caption}) were chosen because of their relatively large object sizes, serving as a counterpart to the small objects in the acne datasets.
For each TexBiG class, we selected two annotators with the largest number of annotated objects to better visualize annotator differences. 


\bibliographystyle{splncs04}
\bibliography{main}